\documentclass[conference]{IEEEtran}
\IEEEoverridecommandlockouts

\usepackage{cite}
\usepackage{amsmath,amssymb,amsfonts}
\usepackage{algorithmic}
\usepackage{graphicx}
\usepackage{textcomp}
\usepackage{xcolor}

\usepackage{threeparttable}
\usepackage{listings}
\usepackage{textcomp}
\usepackage[marginal]{footmisc}
\usepackage{times}
\usepackage{latexsym}
\usepackage[utf8]{inputenc}
\usepackage{microtype}
\usepackage{inconsolata}
\usepackage{array}
\usepackage{pifont}
\usepackage{tabularx}
\usepackage{adjustbox}
\usepackage{multirow}
\usepackage{tablefootnote}
\usepackage{enumerate}
\usepackage{xspace}
\usepackage{tcolorbox}
\usepackage{booktabs,amsfonts,dcolumn}

\usepackage{url}
\usepackage{amsmath,amsthm,amsfonts,amssymb,bm,stmaryrd}

\usepackage{latexsym}
\usepackage{microtype}
\usepackage{soul}
\usepackage{booktabs}
\usepackage{algorithm}
\usepackage{algorithmic}
\usepackage{float}
\usepackage{subfigure}
\usepackage{paralist}
\usepackage{xcolor}
\usepackage{color}
\usepackage{mathtools}
\usepackage{makecell}
\usepackage{multirow}
\usepackage{multicol}
\usepackage{array}
\usepackage{comment}
\usepackage{amsfonts,amssymb}
\usepackage{textcomp}
\usepackage{fmtcount}
\usepackage{flushend}
\usepackage{thmtools}
\usepackage{xspace}
\usepackage{threeparttable}
\usepackage{colortbl}
\definecolor{mygray}{gray}{.9}

\usepackage{balance}

\usepackage{CJKutf8}

\newcommand{\eg}{\hbox{\emph{e.g.}}\xspace}
\newcommand{\etc}{\hbox{\emph{etc}}\xspace}
\newcommand{\ie}{\hbox{\emph{i.e.}}\xspace}
\newcommand{\wrt}{\hbox{\emph{w.r.t.}}\xspace}

\newcommand{\tabincell}[2]{\begin{tabular}{@{}#1@{}}#2\end{tabular}}

\newcommand{\ours}{OpenBG}

\definecolor{highlight}{rgb}{0,0,255} 

\begin{document}

\title{
	 Construction and Applications of Billion-Scale Pre-Trained Multimodal Business Knowledge Graph
}

\author{

\IEEEauthorblockN{Shumin Deng\textsuperscript{$\ast$ $\dagger$}}
\IEEEauthorblockA{
\textit{National University of Singapore} \\
Singapore \\
shumin@nus.edu.sg \\
}

\and

\IEEEauthorblockN{Chengming Wang\textsuperscript{$\ast$}}
\IEEEauthorblockA{
\textit{Alibaba Group} \\
Hangzhou, China \\
chengming.wcm@alibaba-inc.com \\
}

\and 

\IEEEauthorblockN{Zhoubo Li\textsuperscript{$\ast$}}
\IEEEauthorblockA{
\textit{Zhejiang University} \\
Hangzhou, China \\
zhoubo.li@zju.edu.cn \\
}

\and 

\IEEEauthorblockN{Ningyu Zhang\textsuperscript{$\ddagger$}}
\IEEEauthorblockA{
\textit{Zhejiang University} \\
China \\
zhangningyu@zju.edu.cn \\
}

\and

\IEEEauthorblockN{Zelin Dai, Hehong Chen}
\IEEEauthorblockA{
\textit{Alibaba Group} \\
Hangzhou, China \\
zelin.dzl@alibaba-inc.com \\
hehong.chh@alibaba-inc.com \\
}

\and

\IEEEauthorblockN{Feiyu Xiong, Ming Yan}
\IEEEauthorblockA{
\textit{Alibaba Group} \\
Hangzhou, China \\
feiyu.xfy@alibaba-inc.com \\
ym119608@alibaba-inc.com \\
}

\and

\IEEEauthorblockN{Qiang Chen}
\IEEEauthorblockA{
\textit{Alibaba Group} \\
Hangzhou, China \\
lapu.cq@alibaba-inc.com \\
}

\and

\IEEEauthorblockN{Mosha Chen}
\IEEEauthorblockA{
\textit{Alibaba Group} \\
Hangzhou, China \\
chenmosha.cms@alibaba-inc.com \\
}

\and

\IEEEauthorblockN{Jiaoyan Chen}
\IEEEauthorblockA{
\textit{University of Manchester} \\
United Kingdom \\
jiaoyan.chen@manchester.ac.uk \\
}

\and

\IEEEauthorblockN{Jeff Z. Pan}
\IEEEauthorblockA{
\textit{University of Edinburgh} \\
United Kingdom \\
j.z.pan@ed.ac.uk \\
}

\and

\IEEEauthorblockN{Bryan Hooi}
\IEEEauthorblockA{
\textit{National University of Singapore} \\
Singapore \\
dcsbhk@nus.edu.sg \\
}

\and

\IEEEauthorblockN{Huajun Chen\textsuperscript{$\ddagger$}}
\IEEEauthorblockA{
\textit{Zhejiang University} \\
Hangzhou, China \\
huajunsir@zju.edu.cn \\
}

}

\maketitle

\renewcommand\thefootnote{$\ast$}
\footnotetext{Equal contribution.}
\renewcommand\thefootnote{$\dagger$}
\footnotetext{This work was mostly done in Zhejiang University with working as a research intern at Alibaba Group, and completed in NUS.}
\renewcommand\thefootnote{$\ddagger$}
\footnotetext{Corresponding author.}
\renewcommand{\thefootnote}{\arabic{footnote}} 

\begin{abstract}

Business Knowledge Graphs (KGs) are important to many enterprises today, providing factual knowledge and structured data that steer many products and make them more intelligent.
Despite their promising benefits, building business KG necessitates solving prohibitive issues of deficient structure and multiple modalities. 
In this paper, we advance the understanding of the practical challenges related to building KG in non-trivial real-world systems. 
We introduce the process of building an open business knowledge graph (\ours) derived from a well-known enterprise, Alibaba Group. 
Specifically, we define a core ontology to cover various abstract products and consumption demands, with fine-grained taxonomy and multimodal facts in deployed applications. 
{\ours} is an open business KG of unprecedented scale: 2.6 billion triples with more than 88 million entities covering over 1 million core classes/concepts and 2,681 types of relations. 
We release all the open resources ({\ours} benchmarks) derived from it for the community and report experimental results of KG-centric tasks. 
We also run up an online competition based on {\ours} benchmarks, and has attracted thousands of teams. 
We further pre-train {\ours} and apply it to many KG-enhanced downstream tasks in business scenarios, demonstrating the effectiveness of billion-scale multimodal knowledge for e-commerce. 
All the resources with codes have been released at 
\textcolor{blue}{\url{https://github.com/OpenBGBenchmark/OpenBG}}. 

\end{abstract}




\section{Introduction}
\label{sec:intro} 

Knowledge Graphs (KGs) have attracted widespread attention from both academia and industry over the years \cite{J2019_IndustryKG,ACL2021_KGC}, and have been widely applied in many fields, such as education \cite{J2018_KnowEdu,ISWC2021_CKGG}, biomedical science \cite{J2020_MedicalKG,NeurIPS2021_KG4Medical,ICLR2022_OntoProtein}, and finance \cite{COLING2016_KG4Finance,WWW2019KGTA_Event4TSPA}. 
Specifically, business such as e-commerce has increasingly permeated all aspects of our daily life, business KGs \cite{WSDM2020_WalmartKG,ICDE2021_PKGM} have been accumulating a large amount of data in different fields \cite{VLDB2015_DEXTER,AKBC2020_VarSpot,ACL2021_AdaTag}, with the potential to address applications such as recommendation \cite{TKDE2020_KG4Recommender} and product search \cite{TKDE2017_ProductSearchEngines}. 

\textbf{Motivation} \emph{(Why to construct {\ours})}. 
With the tremendous potential of business KG in many applications, we propose to construct a new business KG entitled {\ours}, characterized by 
(1) \emph{standardized organization and integration of business information} about known ‘‘entities'' and ‘‘relations'', as well as hyper-relational knowledge with business axioms and rules, for proprietary resources of an enterprise; 
(2) \emph{utilization of billion-scale multimodal business KG} with pre-training, and serving for various knowledge-enhanced applications.

\begin{figure}[!t]
  \centering
  \includegraphics[width=0.99\linewidth]{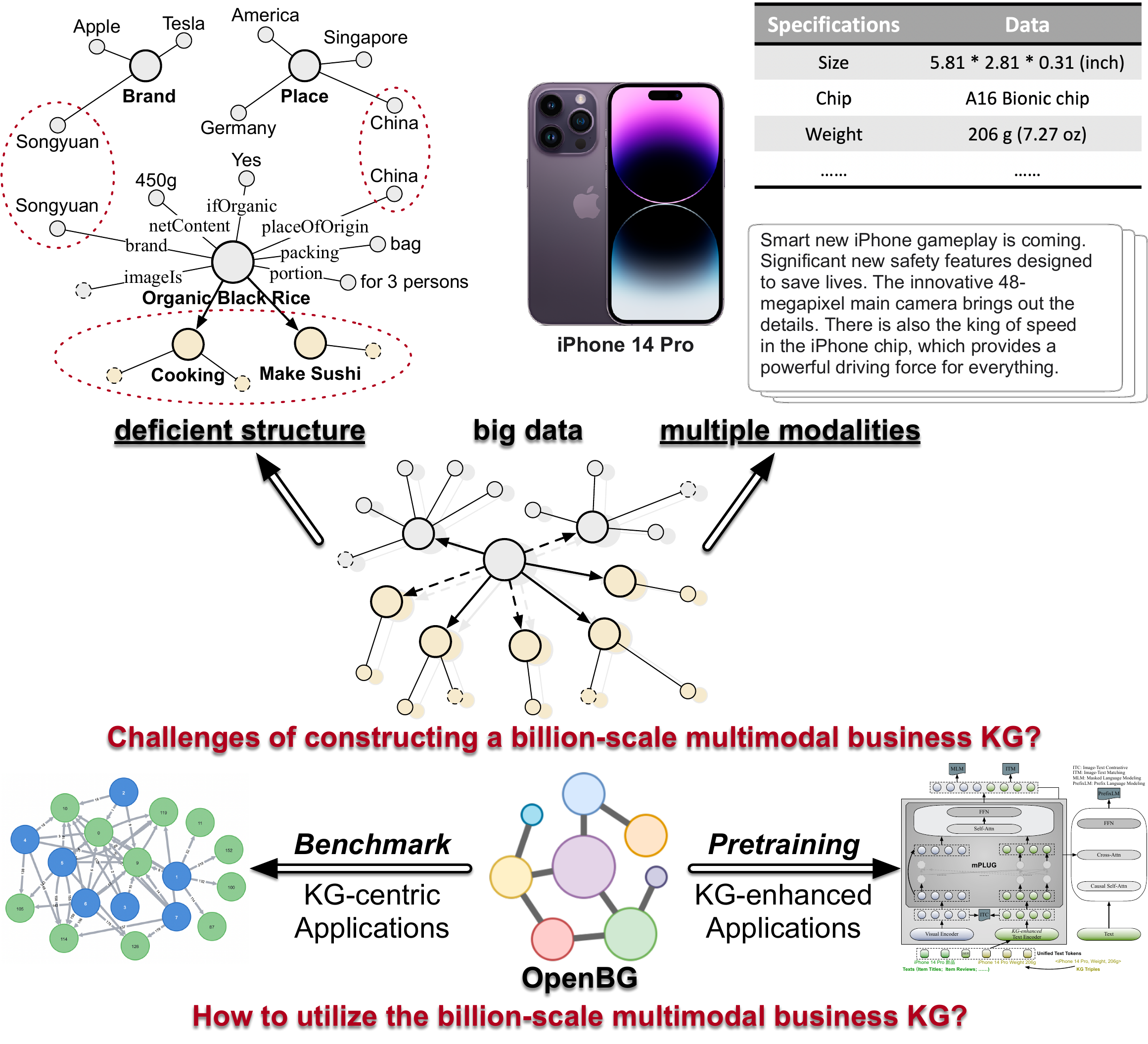}
  \caption{
    Challenges of building a large-scale business KG from multi-source and noisy big data, and illustration of KG applications.
    \label{fig:intro} 
  }
\end{figure}

\textbf{Challenges}. 
{\ours} addresses the following two challenges: 
(1) \emph{Deficient Structure}.
Due to the noise and redundancy that inevitably exist in  big data, large-scale business KGs built with it often suffer from structural defects. For example, as seen in Figure~\ref{fig:intro}, ‘‘China'' is both an instance of the class (Place) and a value of the attribute (\texttt{placeOfOrigin}), showing redundancy in definition; besides, ‘‘Cooking'' and ‘‘Make Sushi'' are closely related via the \texttt{subClassOf} relationship, but not directly linked in the current KG, indicating the lack of completeness. 
(2) \emph{Multiple Modalities}. As big data are derived from multimedia sources, they often possess multiple modalities, \eg, texts, images, and tables, making it challenging to build such a multimodal KG from multi-source data. For example in Figure~\ref{fig:intro}, to present a product ‘‘iPhone 14 Pro'', titles, technical specifications, advertising slogans, and product images all need to be formatted in a universal way.

\textbf{Construction} \emph{(How to construct {\ours})}.
To tackle the challenges and address the limitations of existing technologies, we propose to build a billion-scale \emph{open business knowledge graph} ({\ours}) from multi-source and noisy big data. 
Specifically, we first design a core ontology of {\ours} with references to W3C standards\footnote{\url{https://www.w3.org/}.}, mainly containing 8 types of core classes/concepts and 2,681 types of relations. 
Considering products are the core entities in business scenarios, and often rich in semantics with various attributes, we define the abstract types of products as \underline{\emph{classes}}, entitled \underline{Category} (\begin{CJK*}{UTF8}{gbsn}产品类目\end{CJK*}). 
We also define \underline{Brand} (\begin{CJK*}{UTF8}{gbsn}品牌\end{CJK*}) and \underline{Place} (\begin{CJK*}{UTF8}{gbsn}地点/产地\end{CJK*}) as classes \wrt their rich semantics and close associations with Category. Brand and Place are often described with rich knowledge and almost all products can be linked to them. 

In addition, Category can also establish linkages to other objects without complex semantics, which are regarded as simple classes and defined as \underline{\emph{concepts}} referring to Simple Knowledge Organization System (SKOS)\footnote{\url{https://www.w3.org/TR/skos-primer/}.}. Concepts aim to bridge the semantic gap between user needs and products \cite{SIGMOD2020_AliCoCo,KDD2021_AliCoCo2,KDD2021_AliCG}, which are also vital in business KGs. We define five types of concepts in {\ours} ontology: \underline{Time} (\begin{CJK*}{UTF8}{gbsn}时间\end{CJK*}), \underline{Scene} (\begin{CJK*}{UTF8}{gbsn}场景\end{CJK*}), \underline{Theme} (\begin{CJK*}{UTF8}{gbsn}主题\end{CJK*}), \underline{Crowd} (\begin{CJK*}{UTF8}{gbsn}人群\end{CJK*}), \underline{Market Segment} (\begin{CJK*}{UTF8}{gbsn}细分市场\end{CJK*}). 

We then populate the {\ours} ontology by linking large-scale multimodal product triples to the pre-defined classes/concepts. We also polish the original ontology along with triple accumulation. 
Finally, we release the ongoing {\ours} containing 1,131,579 core classes/concepts, 88,881,723 entities, and 2,603,046,837 triples. 
Since the scale of {\ours} is enormous, we have deployed almost 4,000 persons/day of manpower to annotate the large-scale raw data from open sources, \eg, Alibaba e-commerce platform, and pre-processed the data with 100$\times$V100 GPUs.

\textbf{Applications} \emph{(How to utilize {\ours})}. 
To demonstrate the business value of {\ours}, we apply it in different dimensions. 
(1) \emph{KG-centric Applications.} 
To simplify direct usage of the billion-scale KG, we construct and release {\ours} benchmarks, sampled from {\ours}, for both research analysis and industrial applications. 
We construct three {\ours} benchmarks\footnote{\url{https://github.com/OpenBGBenchmark/OpenBG}.}, including OpenBG-IMG, OpenBG500, and OpenBG500-L. Therein, OpenBG-IMG is a small multimodal KG with 14,718 multimodal entities (27,910 entities in total); OpenBG500 is a single-modal KG with 249,743 entities, and OpenBG500-L is a large-scale version of OpenBG500, containing 2,782,223 entities. We evaluate {\ours} benchmarks with some popular KG embedding models, and the experimental results demonstrate their promising capacity. 
Besides, we also run up an online competition\footnote{\url{https://tianchi.aliyun.com/dataset/dataDetail?dataId=122271&lang=en-us}.} based on {\ours} benchmarks, which has attracted more than 4,700 researchers and thousands of participants (over 3,000 teams) at the time of writing. 

(2) \emph{KG-enhanced Applications.} 
To further unleash the tremendous potential of the billion-scale {\ours}, we pre-train {\ours} and apply it on many downstream tasks in business scenarios, making it serve as a backbone for e-commerce platforms. 
We select several tasks as examples, including link prediction of item categories; NER for item titles; item title summarization; IE for item reviews; salience evaluation for commonsense knowledge. 
We pre-train {\ours} with a vision-language foundation model mPLUG~\cite{EMNLP2022_mPLUG} and other baseline models. 
The experiments reflect that pre-training models enhanced with {\ours} can achieve satisfactory performance on all tasks, especially in low-resource scenarios. 
We also apply the pre-trained {\ours} to provide services on online e-commerce platforms, and present some cases on online systems in this paper. 
Note that during the pre-training stage, we leverage 14$\times$A100 GPUs with 600,000 steps to train the large model for our tasks, and the entire pre-training process costs 32$\times$V100 GPUs for 120 days approximately.

\textbf{Contributions} of this work can be summarized: 
\begin{itemize}

  \item We introduce a detailed process of constructing a billion-scale multimodal pre-trained business KG ({\ours}), with standard KG schema definitions, which have been applied to real-world systems, hopefully enlightening KG construction in other fields. 

  \item We publish {\ours}, an ongoing business KG, containing more than 88 million entities covering over 1.1 million classes/concepts and over 2.6 billion triples. 
  To the best of our knowledge, {\ours} is the most comprehensive business KG publicly available on the Web. 

  \item We release {\ours} benchmarks and run an online competition for KG-centric applications through {\ours} pre-training, which have attracted thousands of participants, and we also integrate {\ours} to large pre-trained models for KG-enhanced applications. 

\end{itemize}

The remainder of this paper is organized as follows. 
Sec.\ref{sec:practice} shows the process of building billion-scale \ours. 
Sec.\ref{sec:resources} introduces construction and evaluation of the released {\ours} benchmarks. 
Sec.\ref{sec:application} presents some KG-centric and KG-enhanced applications with pre-trained {\ours}.
Sec.\ref{sec:related_work} reviews related work on business KGs. 
Sec.\ref{sec:con_fw} concludes and discusses the future work.


\section{{\ours} Construction} 
\label{sec:practice} 

\subsection{Preliminary and Ontology Overview}

\begin{figure}[!htbp]
  \centering
  \includegraphics[width=0.99\linewidth]{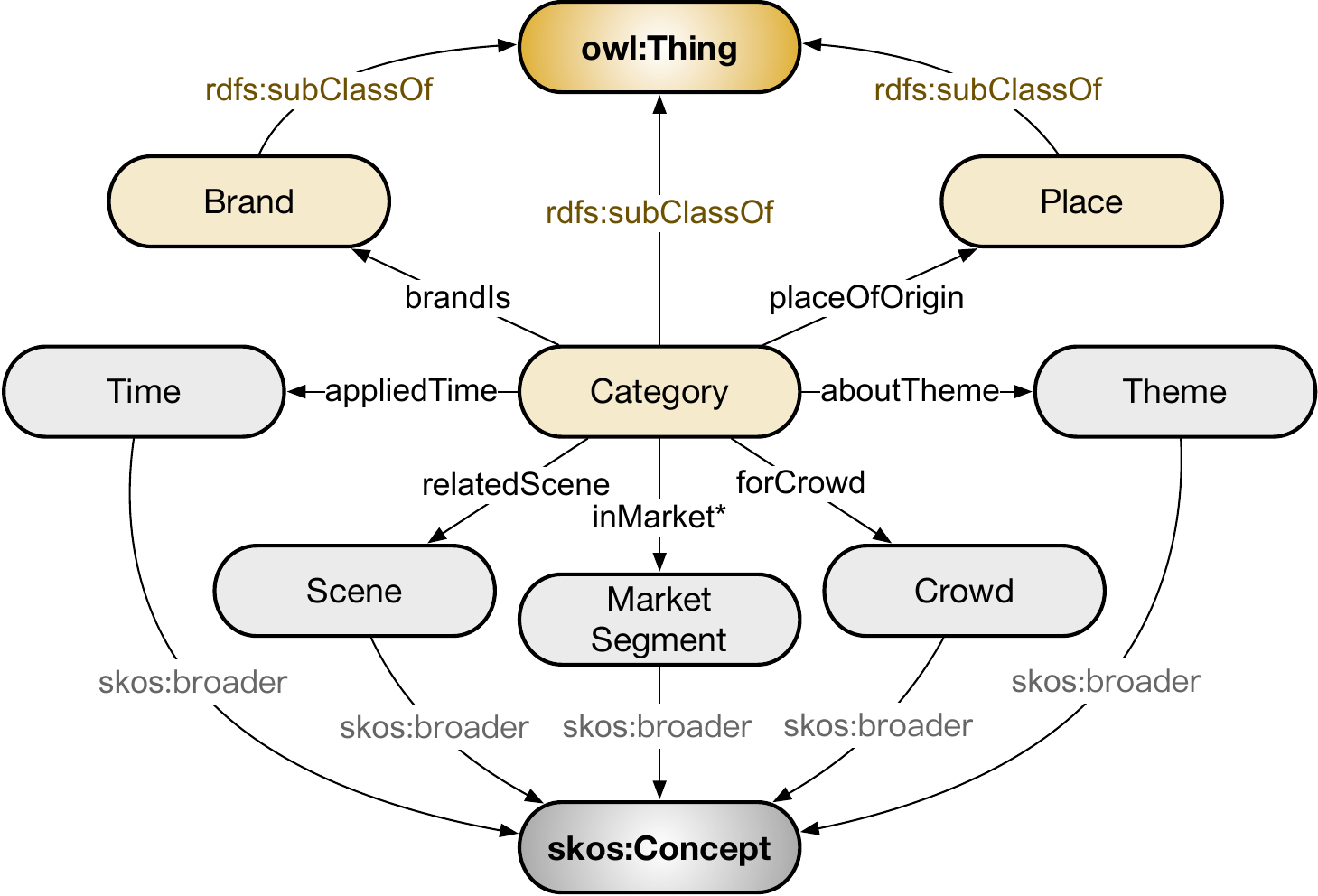}
  \caption{
  The core ontology of \ours. 
  Note that nodes marked in \textcolor[RGB]{222,173,38}{yellow} denotes \textcolor[RGB]{222,173,38}{\emph{classes}} which are subclasses of \texttt{owl:Thing} (linked with \texttt{rdfs:subClassOf}), and nodes marked in \textcolor[RGB]{	105,105,105}{gray} denotes \textcolor[RGB]{105,105,105}{\emph{concepts (simple classes)}} which are subclasses of \texttt{skos:Concept} (linked with \texttt{skos:broader}). 
  inMarket$^*$ denotes a set of relations between Category and Market Segment, represented in abstract form due to space limitations. 
  \label{fig:ontology} 
  }
\end{figure}

To address the challenge of \emph{deficient structure} in business KG construction, we propose to define a canonical ontology or schema) for it with references to W3C standards\footnotemark[1], which imposes constraints on the links coupled with business logic. Furthermore, as the scale of business data is often enormous and unmanageable, defining the standard ontology is fundamental and necessary for business KG construction. 
We present the core ontology of {\ours} in Figure~\ref{fig:ontology}, mainly containing \textcolor[RGB]{222,173,38}{classes}/\textcolor[RGB]{105,105,105}{concepts} and relations. Note that we focus on ontology definition and KG construction in this section, and omit technical details about raw knowledge acquisition (mostly general NLP methods, such as named entity recognition, relation extraction, and coreference resolution which are already resolved in data pre-processing) due to page limits. 
As {\ours} is billion-scale and the raw data are also enormous, the data processing is time-consuming and labor-intensive. 
\emph{The annotation of raw data has cost 4,000 person/day approximately.}
\emph{We deploy the model on 100$\times$V100 GPUs for daily new data ingestion and processing.}
\emph{We also carry out routine maintenance of {\ours} quality, which costs almost 30 person/day.}

\textbf{Automation of {\ours}.}  
Given  the defined ontology and pre-processed structured data, we utilize a free and open source Java framework called Apache Jena\footnote{\url{https://jena.apache.org/}.} to automatically 
(1) formalize {\ours} ontology with Jena ontology API, and then 
(2) populate {\ours} ontology by linking instances to it with RDF (Resource Description Framework) API.

\textbf{Symbol Overview}: 
The {\ours} ontology $\mathbb{O} = \{ \mathbb{C}, \mathbb{P}, \mathcal{R} \}$ mainly comprises \textcolor[RGB]{222,173,38}{classes} $\mathbb{C}$, \textcolor[RGB]{105,105,105}{concepts} $\mathbb{P}$, and \emph{relations} $\mathcal{R}$. 
Therein,
$\mathbb{C} = \mathbb{C}^C \cup \mathbb{C}^B \cup \mathbb{C}^P$ contains \textcolor[RGB]{222,173,38}{Category} $\mathbb{C}^C$, \textcolor[RGB]{222,173,38}{Brand} $\mathbb{C}^B$, and \textcolor[RGB]{222,173,38}{Place} $\mathbb{C}^P$. 
A category $c \in \mathbb{C}^C$ is regarded as the abstract types of a \emph{product} $u$, denoted by $\langle u, \texttt{rdf:type}, c \rangle$, where \emph{products} are core entities in {\ours}. Category $\mathbb{C}^C$ with a rich hierarchy is defined to indicate the taxonomy of \emph{product}, and Brand $\mathbb{C}^B$, Place $\mathbb{C}^P$ as well as concepts $\mathbb{P}$ are all linkable to it via relations $\mathcal{R}$.

\textbf{Core Classes/Concepts}: 
As seen in Figure~\ref{fig:ontology}, the central node in {\ours} ontology is \textcolor[RGB]{222,173,38}{\underline{Category}}, which indicates the taxonomy of \textcolor[RGB]{222,173,38}{\emph{products}} with various attributes, such as size, weight, and color. Considering products are informative with rich semantics, we define the \textcolor[RGB]{222,173,38}{\emph{abstract types of products}} (named \textcolor[RGB]{222,173,38}{\underline{Category}}) as \textcolor[RGB]{222,173,38}{\underline{\textbf{classes}}} in the ontology. 
For example, mobile phone, as an abstract type of iPhone 14 Pro, can be defined as a class of Category. 
In addition, products can be linkable to other objects, such as \textcolor[RGB]{222,173,38}{brands} and \textcolor[RGB]{222,173,38}{places}. We also define their abstract types as  subclasses of  \textcolor[RGB]{222,173,38}{\underline{Brand}} and \textcolor[RGB]{222,173,38}{\underline{Place}} class \wrt their rich knowledge, \eg, a brand also has various attributes such as founder and logo. Furthermore, products can also establish relations to brands and places. 

Besides, products also can be linked to some objects without complex semantics, but these objects can create a bridge between products and users' needs. For example, iPhone 14 Pro is related to a scenario of \emph{giving gifts} (not requiring rich-semantic description), which represents the business knowledge that iPhone 14 Pro is commonly given as a gift. 
We define these simple-semantic objects as \textcolor[RGB]{105,105,105}{\textbf{concepts}}, inspired by AliCoCo~\cite{SIGMOD2020_AliCoCo} (an e-commerce cognitive concept net). 
Concepts are deemed as simple classes, because they are abstract entities without complex semantics. We also refer to Simple Knowledge Organization System (SKOS)\footnotemark[2] to standardize concept definitions. 

Overall, we define \emph{3 types of core \textcolor[RGB]{222,173,38}{classes}} to express knowledge related to products (\ie, ‘‘\textcolor[RGB]{222,173,38}{\underline{Category}}'', ‘‘\textcolor[RGB]{222,173,38}{\underline{Place}}'', ‘‘\textcolor[RGB]{222,173,38}{\underline{Brand}}'') and \emph{5 types of core \textcolor[RGB]{105,105,105}{concepts}} to express knowledge related to business concepts (\ie, ‘‘\textcolor[RGB]{105,105,105}{\underline{Time}}'', ‘‘\textcolor[RGB]{105,105,105}{\underline{Scene}}'', ‘‘\textcolor[RGB]{105,105,105}{\underline{Market Segment}}'', ``\textcolor[RGB]{105,105,105}{\underline{Crowd}}'' and ``\textcolor[RGB]{105,105,105}{\underline{Theme}}'').

\textbf{Core Relations}: 
$\mathcal{R} = \{ \mathcal{R}_{obj}, \mathcal{R}_{meta}, \mathcal{R}_{data} \}$ includes three types of relations, including object properties $\mathcal{R}_{obj}$, meta-properties $\mathcal{R}_{meta}$, and data properties $\mathcal{R}_{data}$. 

(1) $\mathcal{R}_{obj}$: Object properties reflect relationships among classes/concepts, which constrain the type of head entity (domain) and tail entity (range). For example, \texttt{placeOfOrigin} conveys association between Category $\mathbb{C}^C$ and Place $\mathbb{C}^P$, thus the domain of \texttt{placeOfOrigin} must belongs to $\mathbb{C}^C$, and the range must belongs to $\mathbb{C}^P$. 
As seen in Figure~\ref{fig:ontology}, we define some \emph{object properties} to model associative relationships between Category and other core classes/concepts, such as \texttt{brandIs}, \texttt{placeOfOrigin}, \texttt{appliedTime}, \texttt{relatedScene}, \texttt{aboutTheme}, \texttt{forCrowd}, \texttt{inMarket$^*$} (a set of relations, in short for brevity). 

(2) $\mathcal{R}_{meta}$: Meta-properties (also called built-in properties) indicate axioms of the ontology defined by W3C standards\footnotemark[1]. 
We import four types of \emph{meta-properties} to express taxonomy (\texttt{rdfs:subClassOf}, \texttt{skos:broader}), synonymy (\texttt{owl:equivalentClass}) and instantiation (\texttt{rdf:type}) of classes/concepts. 
For example, $\langle$Category, \texttt{rdfs:subClassOf}, \texttt{owl:Thing}$\rangle$, $\langle$Scene, \texttt{skos:broader}, \text{skos:Concept}$\rangle$, and $\langle c/p$, \texttt{owl:equivalentClass}, $x \rangle$, where $c \in \mathbb{C}$, $p \in \mathbb{P}$, and $x$ denotes an accessible exogenous object linked to {\ours}. 
Moreover, we also import two other types of \emph{meta-properties} to express properties of properties, \ie, \texttt{rdfs:subPropertyOf} and \texttt{owl:equivalentPropertyOf}.

(3) $\mathcal{R}_{data}$: Data properties are used to describe the attributes of classes. 
For example, the iPhone 14 Pro product has the data properties of size, weight, color, and so on. 
Considering that data properties in business scenarios are often derived from the general domain, we also link some data properties in {\ours} to cnSchema\footnote{\url{https://github.com/cnschema/cnSchema}.} (an open Chinese KG schema, application-oriented) with two meta-properties: \texttt{rdfs:subPropertyOf} and \texttt{owl:equivalentPropertyOf}.

\begin{figure*}[!htbp]
  \centering
  \includegraphics[width=1.0\linewidth]{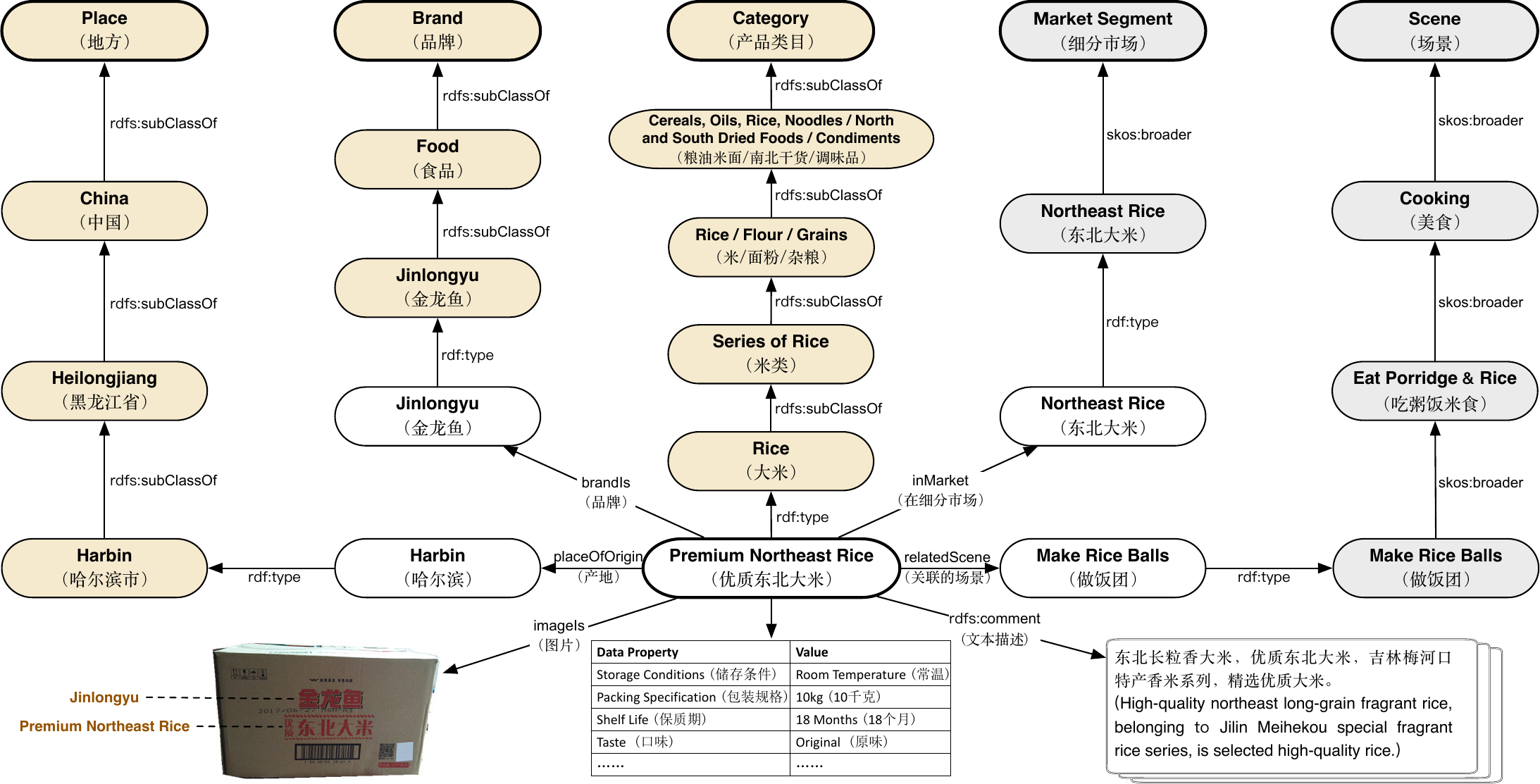}
  \caption{A snapshot of \ours. 
  The rounded rectangles with no color fill denote \emph{instances} of \textcolor[RGB]{222,173,38}{classes}/\textcolor[RGB]{105,105,105}{concepts}. 
  \label{fig:exp_example} }
\end{figure*}

The detailed statistics of the ongoing {\ours} in current version are shown in Table~\ref{tab:openbg-full_stat}. 
We also show a snapshot of {\ours} containing some examples in Figure~\ref{fig:exp_example}. 

\begin{table}[!htbp]
\centering
\small


\caption{
	\label{tab:openbg-full_stat} 
	Statistics of {\ours} at the time of writing. 
}

\begin{tabular}{p{5.32cm} | p{2.52cm}}
\toprule
\rowcolor{mygray}\multicolumn{2}{l}{\textbf{Overall}} \\
\midrule

\# core classes 				& \footnotesize{460,805} \\
\# core concepts 				& \footnotesize{670,774} \\
\# relation types 						& \footnotesize{2,681} \\
\# products (instances of categories) 	& \footnotesize{3,062,313} \\ 
\# triples 								& \footnotesize{2,603,046,837} \\

\bottomrule
\end{tabular}

\vspace{0.5mm}

\resizebox{\linewidth}{!}{
\begin{tabular}{c | c c c c c c}
\toprule

\rowcolor{mygray}\tabincell{c}{ \textbf{Core Class/}\\ \textbf{Concept} } & 
\textbf{\# level1} & 
\textbf{\# level2} & 
\textbf{\# level3} & 
\textbf{\# level4} & 
\textbf{\# level5} & 
\enspace \textbf{\# All} / \textbf{\# leaf} \\

\midrule

Category 		& 93 	& 889 	& 3,069 & 3,049 & / 	& 7,100 	/ 3,399 \\
Brand 			& 45 	& 411234 & / & / & / & 411,279 / 411,234 \\
Place 			& 208 	& 266 	& 333 	& 2,847 &38,773 & 42,426 	/ 38,969 \\
Scene 			& 19 	& 4,027 & 617 	& 729 	& / 	& 5,392 	/ 5,170 \\
Crowd 			& 8 	& 37 	&45,105 & 57 	& / 	& 45,207 	/ 45,180 \\
Theme 			& 14 	& 5,219 & 143 	& 143 	& / 	& 5,519 	/ 5,360 \\
Time 			& 3 	& 55 	& / 	& / 	& / 	& 58 		/ 55 \\
Market\_S 		& 614,598 & / 	& / 	& / 	& / 	& 614,598 	/ 614,598 \\

\bottomrule
\end{tabular}
}

\vspace{0.5mm}

\resizebox{\linewidth}{!}{
\begin{tabular}{c | c | c | c | c}
\toprule
\rowcolor{mygray}\multicolumn{5}{l}{\textbf{Core Relation}} \\
\midrule

\multirow{4}*{ \tabincell{c}{\textbf{object} \\ \textbf{property}} } & 

\# brandIs 	& \# placeOfOrigin  & \# relatedScene 	& \# forCrowd \\
& 1,900,673 	& 2,155,721 		& 28,766,037 		& 3,966,635 \\
\cmidrule{2-5}

& \# aboutTheme & \# appliedTime & \# inMarket$^*$ &  \\
& 132,135 		& 773,371 		 & 1,654,057,075 &  \\

\midrule

\multirow{4}*{ \tabincell{c}{\textbf{data} \\ \textbf{property}} } & 

\# rdfs:label 	& \# labelEn & \# skos:prefLabel & \# skos:altLabel \\
& 3,062,313 	& 3,072,337  & 670,774 			 & 670,774 \\

\cmidrule{2-5}

& \# rdfs:comment & \# imageIs & \multicolumn{2}{c}{\# product attributes} \\ 
& 3,062,300 & 102,504 & \multicolumn{2}{c}{75,486,855} \\

\midrule

\multirow{4}*{ \tabincell{c}{\textbf{meta-} \\ \textbf{property}} } & 
\# rdfs:subClassOf & \# skos:broader & \# rdf:type & \# owl:equivalentClass \\
& 460,760 & 670,774 & 88,881,723 & 496,086 \\

\cmidrule{2-5}

& \multicolumn{2}{c|}{\# rdfs:subPropertyOf} & \multicolumn{2}{c}{\# owl:equivalentPropertyOf} \\
& \multicolumn{2}{c|}{1,018} & \multicolumn{2}{c}{2,448} \\

\bottomrule
\end{tabular}
}

\vspace{-1.2mm}
\end{table}

\subsection{Construction of Ontology Classes} \label{sec:method_class}

In this section, we present the construction of 3 core classes and their subclasses, \ie, ‘‘Category'', ‘‘Place'', ‘‘Brand'', which lie in the $2_{nd}$ level of the ontology (just below \texttt{owl:Thing}).

\textbf{Construction of ‘‘Category''.} 
We follow \emph{a top-down approach} to constructing Category. 

(1) \emph{Define ‘‘Category'' and Taxonomy}. 
We first create a single top-level class: Category, before specializing it with subclasses. We further break down each class layer by layer, for example, \emph{electronic components} are broken down into some subclasses, such as \emph{LED}, \emph{power supply}, \etc. 
To evaluate the quality of Category definition, we schedule qualified human resources (experts in e-commerce, 30 person/day) to go through a daily review process.
The main concern of Category quality includes: 
(i) if the class label definition is clear and reasonable enough; 
(ii) if a category contains its child nodes completely; 
(iii) if a category excludes child nodes of other categories; 
(iv) if products associated with the category category is popular enough;
(v) if the category is acknowledged by customers and sellers. 
Domain experts will rate and tune each category according to these factors every day.

(2) \emph{Create Multimodal Instances of ‘‘Category''}. 
In fact, the instances of ‘‘Category'' are products in business scenarios. We randomly sample some products for each leaf node in category taxonomy from Alibaba e-commerce platform. 
To address the challenge of \emph{multiple modalities} in business KG construction, we formalize multimodal information of products all in triple format with different relations. 
To organize \emph{semi-structured association and attribute information}, we associate products with other classes/concepts through object properties $\mathcal{R}_{obj}$, \eg, $\langle$iPhone 14 Pro, \texttt{brandIs}, Apple$\rangle$;  
we then define data properties $\mathcal{R}_{data}$ of each product to express its attribute-related information, \eg, $\langle$iPhone 14 Pro, \texttt{weight}, 206g$\rangle$. 
Particularly, to organize \emph{unstructured attribute information}, 
we retrieve the textual corpus of product description and images of products from Alibaba e-commerce platform, and link them to products with two data properties (\texttt{rdfs:comment}, \texttt{imageIs}), enabling \textbf{\emph{multimodal knowledge about products}}, \ie. textual descriptions and images.

\textbf{Construction of ‘‘Place'' and ‘‘Brand''.} 
We generally utilize \emph{a schema mapping approach} \cite{ISWC2021_CKGG,CCKS2017_CrowdGeoKG} to construct ‘‘Place'' and ‘‘Brand'' from diverse sources. 

(1) \emph{Define ‘‘Place'' and Taxonomy}. 
We integrate and transform administrative region data in different formats from diverse KGs, such as Wikidata\footnote{\url{https://www.wikidata.org/}.} and OpenKG\footnote{\url{http://openkg.cn/dataset/xzqh}.}. 
The taxonomy of ‘‘Place'' includes country or great region, province, city, county, village/town in order. 

(2) \emph{Define ‘‘Brand'' and Taxonomy}. 
‘‘Brand'' is divided into 45 major categories following the guideline for declaration of goods\footnote{\url{http://sbj.cnipa.gov.cn/sbsq/sphfwfl/}.}, \eg, Clothes, Furniture, and Vehicle. We also integrate and transform various kinds of brand-related data from authoritative websites, \eg, the logo of a brand on its official homepage. 

(3) \emph{Link ‘‘Place'' and ‘‘Brand'' with Products}. 
We link ‘‘Place'' and ‘‘Brand'' to products via matching their textual labels. 
Concretely, for each product containing place and brand information, we map the textual labels of its place and brand to standard names defined in ‘‘Place'' and ‘‘Brand'' taxonomy, 
by jointly conducting trie prefix tree precise matching and fuzzy matching of synonyms.

\subsection{Construction of Ontology Concepts} \label{sec:method_concept}
{\ours} currently contains five types of core concepts, \ie, Scene, Crowd, Theme, Time, and Market Segment. 
We follow \emph{a bottom-up approach} for Concept taxonomy construction. 

(1) \emph{Create Instances of ‘‘Concept''}. 
We first leverage sequence labeling to extract concepts from real-world large-scale business text corpus, such as user-written reviews, product titles, and search queries.
We utilize BERT-CRF model, a popular model for many sequence labeling tasks \cite{EMNLP2020_OpenUE,SIGIR@BIRNDL2019_BERT-CRF}. 
BERT-CRF model consists of a BERT layer and a CRF (Conditional Random Field) layer, where BERT \cite{NAACL2019_BERT} enables to obtain a contextual representation for each word and CRF considers the correlations between the current label and neighboring labels. 

(2) \emph{Define ‘‘Concept'' and Taxonomy}. 
Given the five top-level broad concepts: Scene, Crowd, Theme, Time, and Market Segment, we classify instances of ‘‘Concept'' into these five pre-defined concepts. Then we summarize the narrower concepts to broader ones level by level, until to the top level. 
Moreover, to evaluate the quality of concepts, we formulate concepts in leaf nodes following multi-faceted commonsense knowledge \cite{AKBC2020_Commonsense}, mainly including four dimensions: 

(i) \emph{Plausibility}: Indicating whether a concept-oriented statement is meaningful, \eg, $\langle$sports shoes, \texttt{forCrowd}, the elderly$\rangle$ where sports shoes are products and the elderly are a narrower concept of Crowd, is a meaningful statement for ‘‘the elderly''. 

(ii) \emph{Typicality}: Indicating whether a concept-oriented statement is valid for the majority of instances, \eg, $\langle$lightweight sports shoes, \texttt{forCrowd}, the elderly$\rangle$ are typical while $\langle$trendy sports shoes, \texttt{forCrowd}, the elderly$\rangle$ are not, because trendy shoes are not suitable for most of elderly people. 

(iii) \emph{Remarkability}: Indicating whether a concept is distinguishable enough from closely related ones, \eg, $\langle$non-slip shoes, \texttt{forCrowd}, the elderly$\rangle$ are remarkable while $\langle$thin and light shoes, \texttt{forCrowd}, the elderly$\rangle$ are not, because thin and light shoes are also suitable for young people. 

(iv) \emph{Salience}: Indicating whether a concept is representative enough, so that instances can be associated with it spontaneously. 
Generally, a statement both satisfying \emph{Typicality} and \emph{Remarkability} implies \emph{Salience}. 


\section{{\ours} Benchmark: KG-centric Application}
\label{sec:resources}

To promote further research in developing plausible KG representation solutions to real-world applications, we present the {\ours} Benchmark, a challenging benchmark to facilitate reproducible, scalable, and multimodal KG research. 
{\ours} Benchmark is sampled from {\ours}, covering most of the features defined in {\ours}. 
Specifically, {\ours} Benchmark is designed following three principles:
(1) Key problems in business KG. We select tasks to address crucial academic and industrial issues that remain to be solved for business KG.
(2) Business application-centric task design. All the tasks in the benchmark are selected based on business scenarios in Alibaba Group to address real-world applications.
(3) Quality control in long-term maintenance. We utilize several denoising strategies and artificially review data to ensure data quality. 
Thus, {\ours} Benchmark is mature enough to be considered for industry-wide acceptance. 
We also conduct experiments\footnotemark[3] and release the leaderboard\footnotemark[4] publicly. 

\subsection{Benchmark Construction} 

{\ours} benchmark contains several subsets, including OpenBG-IMG, OpenBG500, and OpenBG500-L. 
Therein, OpenBG-IMG is a small multimodal dataset, OpenBG500 is a single-modal dataset, while OpenBG500-L is a large-scale version of OpenBG500, aiming to benchmark efficient machine learning methods over the large-scale KG.
We show the statistical details of the {\ours} benchmark with comparison to some existing large-scale multimodal KG Wikidata5M~\cite{J2021_KEPLER} and OGB~\cite{NeurIPS2020_OGB,NeurIPS2021_OGB-LSC} in Table~\ref{tab:OpenBG_stat}. 

\begin{table}[!htbp]
\centering
\small

\caption{
	Summary statistics of OpenBG datasets. 
	\emph{Note that OpenBG (Full) has no train/dev/test split, and 
	OGB-LSC refers to WikiKG90Mv2 in OGB-LSC. 
	$\star$: There are 14,718 multi-modal entities in OpenBG-IMG.} 
	\label{tab:OpenBG_stat}
}

\resizebox{\linewidth}{!}{
\begin{tabular}{l r r r r r}

	\toprule

	\multicolumn{1}{c}{\textbf{Dataset}} & 
	\multicolumn{1}{c}{\textbf{\# Ent}} & 
	\multicolumn{1}{c}{\textbf{\# Rel}} & 
	\multicolumn{1}{c}{\textbf{\# Train}} & 
	\multicolumn{1}{c}{\textbf{\# Dev}} & 
	\multicolumn{1}{c}{\textbf{\# Test}} \\
	
	\midrule

	OpenBG-IMG 	& 27,910$^\star$   & 136    & 230,087   & 5,000  & 14,675  \\
	OpenBG500 	& 249,743  & 500   & 1,242,550  & 5,000 & 5,000  \\ 
	OpenBG500-L & 2,782,223 & 500   & 47,410,032  & 10,000  & 10,000 \\
	OpenBG (Full)  & 88,881,723 & 2,681 &  260,304,683 & - & - \\
	
	\midrule

	Wikidata5M  & 4,594,485  & 822   & 20,614,279  & 5,163   &  5,133 \\
	OGB-LSC  & 91,230,610  & 1,387  & 608,062,811  &  15,000  &  10,000 \\

	\bottomrule

\end{tabular}
}

\end{table}

\begin{figure}[!htbp]
  \centering
  \includegraphics[width=0.9\linewidth]{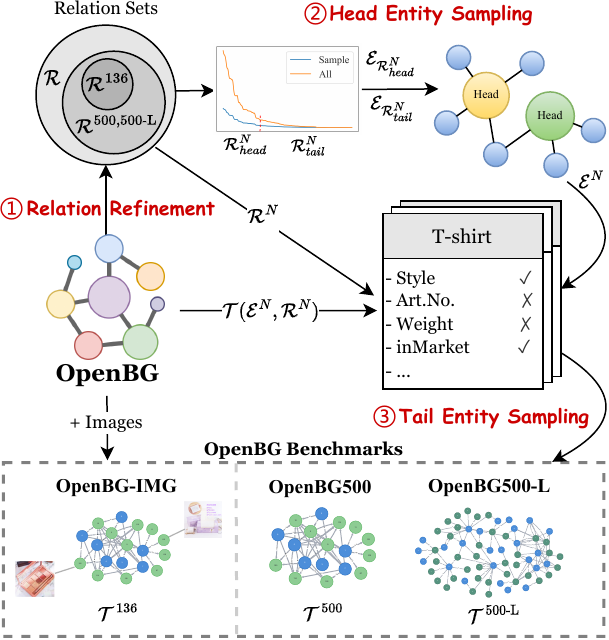}
  \caption{Building process of {\ours} benchmarks.  
  \label{fig:openbg_benchmark_workflow} }
\end{figure}

Given \ours~(full) with a full set of entities, relations, triples, denoted by $\{ \mathcal{E}, \mathcal{R}, \mathcal{T} \}$, we utilize a three-stage method to build high-quality {\ours} benchmarks, including \textbf{step 1}: selecting relations  from $\mathcal{R}$ (\emph{relation refinement}); \textbf{step 2}: filtering head entities from $\mathcal{E}$ (\emph{head entity filtering}); \textbf{step 3}: sampling tail entities in triples $\mathcal{T}$ (\emph{tail entity sampling}). 
We illustrate the process of building {\ours} benchmarks in Figure~\ref{fig:openbg_benchmark_workflow}. 

\begin{figure}[!htbp]
  \centering
  \includegraphics[width=0.94\linewidth]{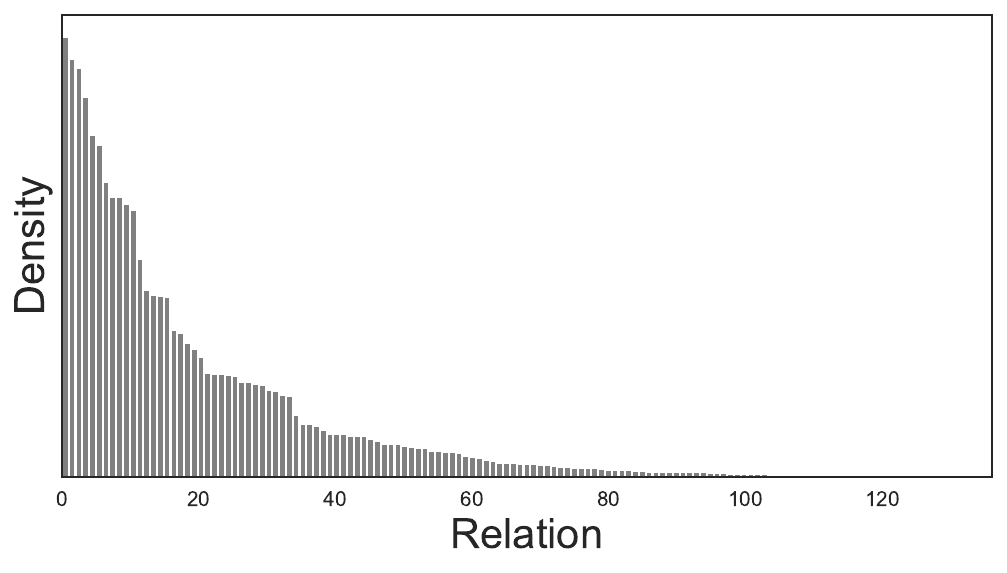}
  \caption{
    Relation distribution of OpenBG-IMG.
    \label{fig:relation_distribution}
  }
\end{figure}

(1) \emph{Relation Refinement.} 
We follow two principles to manually filter representative relations in \ours: 
(i) To select \emph{high-frequency relations}, as they are supposed to imply a relatively high-value applications. 
(ii) To select \emph{closely business-related relations}, as they directly describe the attributes of products. 
In this way, we obtain relation subsets $\mathcal{R}^{\text{500}}, \mathcal{R}^{\text{500-L}}$ and $\mathcal{R}^{\text{136}}$, respectively containing 500 (for OpenBG500 \& OpenBG500-L) and 136 (for OpenBG-IMG) relations, where $\mathcal{R}^{\text{136}} \subset \mathcal{R}^{\text{500}}, \mathcal{R}^{\text{500-L}} \subset \mathcal{R}$. Note that some triples contain no image, thus OpenBG-IMG has fewer relations than OpenBG500. 
We show the distribution of 136 relations in OpenBG-IMG, as seen in Figure~\ref{fig:relation_distribution}, which follows a long-tail distribution. 
OpenBG500 as well as OpenBG500-L also follow similar relation distributions, and we omit figures due to page limits. 

(2) \emph{Head Entity Filtering.} 
Regarding the long-tail relation distribution of $\mathcal{R}^{\text{136}}$,  $\mathcal{R}^{\text{500}}$, and $\mathcal{R}^{\text{500-L}}$, we split $\mathcal{R}^N$ ($N$ = 136, 500, 500-L)  into \emph{head-relations} $\mathcal{R}^N_{head}$ and \emph{tail-relations} $\mathcal{R}^N_{tail}$. 
Then we divide the entities into $\mathcal{R}^N_{head}$ entities ($\mathcal{E}_{\mathcal{R}^N_{head}}$) and $\mathcal{R}^N_{tail}$ entities ($\mathcal{E}_{\mathcal{R}^N_{tail}}$), and also distribute a sampling rate $\alpha^N_h$ for the former, as well as $\alpha^N_l$ for the latter, where $\alpha^N_h > \alpha^N_l$.
The entity sampling can be denoted by: 
\begin{equation}
\mathcal{E}^N = Sample(\mathcal{E}_{\mathcal{R}^N_{head}},\alpha^N_h) + Sample(\mathcal{E}_{\mathcal{R}^N_{tail}},\alpha^N_l)
\end{equation} 
where $Sample(\mathcal{S}, \alpha)$ denotes randomly sampling the set $\mathcal{S}$ at a certain rate $\alpha$, and 
$\mathcal{E}^N$ (where $N$ = 136, 500, 500-L) is the sampled head entities for triples.

(3) \emph{Tail Entity Sampling.} 
Given the filtered relations $\mathcal{R}^N$ ($N$ = 136, 500, 500-L) and sampled head entities $\mathcal{E}^N$ in triples, 
we firstly filter triples $\mathcal{T}(\mathcal{E}^N, \mathcal{R}^N)$ from $\mathcal{T}$, with head entities in $\mathcal{E}^N$, relations in $\mathcal{R}^N$ and all corresponding tail entities of the given relations. Then we build three benchmarks $\mathcal{T}^N$ ($N$ = 136, 500, 500-L) entitled OpenBG-IMG, OpenBG500, OpenBG500-L through sampling tail entities from  $\mathcal{T}(\mathcal{E}^N, \mathcal{R}^N)$ and connecting them with head entities $\mathcal{E}^N$ and relations $ \mathcal{R}^N$ to obtain the whole triples, denoted by:
\begin{equation}
\mathcal{T}^N = Sample(\mathcal{T}(\mathcal{E}^N,\mathcal{R}^N),\alpha^N)
\end{equation}
where $\alpha^N$ is determined by each benchmark $\mathcal{T}^N$.

\begin{table*}[!htbp]
  \centering
  \small

\caption{\label{tab:openbg-img}
  Results of link prediction on \textbf{OpenBG-IMG}.
  The bold numbers denote the best results.
}
  
\begin{tabular}{l | c c c c c}
\toprule

  \multirow{1}{*}{Model} 
  & \multicolumn{1}{c}{Hits@1 $\uparrow$} 
  & \multicolumn{1}{c}{Hits@3 $\uparrow$} 
  & \multicolumn{1}{c}{Hits@10 $\uparrow$} 
  & \multicolumn{1}{c}{MR $\downarrow$}   
  & \multicolumn{1}{c}{MRR $\uparrow$} \\

\midrule
  \multicolumn{6}{c}{\textit{Single-Modal Approaches}} \\
\midrule

  \multicolumn{1}{l|}{TransE~\cite{NeurIPS2013_TransE}} 
  & \multicolumn{1}{c}{0.150} 
  & \multicolumn{1}{c}{0.387} 
  & \multicolumn{1}{c}{0.647} 
  & \multicolumn{1}{c}{118} 
  & \multicolumn{1}{c}{0.315} \\

  \multicolumn{1}{l|}{TransH~\cite{AAAI2014_TransH}} 
  & \multicolumn{1}{c}{0.129} 
  & \multicolumn{1}{c}{0.525} 
  & \multicolumn{1}{c}{0.743} 
  & \multicolumn{1}{c}{112} 
  & \multicolumn{1}{c}{0.357} \\

  \multicolumn{1}{l|}{TransD~\cite{ACL2015_TransD}} 
  & \multicolumn{1}{c}{0.137} 
  & \multicolumn{1}{c}{0.532} 
  & \multicolumn{1}{c}{0.746} 
  & \multicolumn{1}{c}{110} 
  & \multicolumn{1}{c}{0.364} \\
    
  \multicolumn{1}{l|}{DistMult~\cite{ICLR2015_DistMult}}
  & \multicolumn{1}{c}{0.060} 
  & \multicolumn{1}{c}{0.157} 
  & \multicolumn{1}{c}{0.279} 
  & \multicolumn{1}{c}{524} 
  & \multicolumn{1}{c}{0.139} \\
   
  \multicolumn{1}{l|}{ComplEx~\cite{ICML2016_ComplEx}}
  & \multicolumn{1}{c}{0.143} 
  & \multicolumn{1}{c}{0.244} 
  & \multicolumn{1}{c}{0.371} 
  & \multicolumn{1}{c}{782} 
  & \multicolumn{1}{c}{0.221} \\

  \multicolumn{1}{l|}{TuckER~\cite{ACL2019_TuckER}}
  & \multicolumn{1}{c}{\textbf{0.497}} 
  & \multicolumn{1}{c}{\textbf{0.690}} 
  & \multicolumn{1}{c}{0.820} 
  & \multicolumn{1}{c}{1473} 
  & \multicolumn{1}{c}{\textbf{0.611}} \\
    
   

    
  \multicolumn{1}{l|}{KG-BERT~\cite{2019_KG-BERT}}
  & \multicolumn{1}{c}{0.092} 
  & \multicolumn{1}{c}{0.207} 
  & \multicolumn{1}{c}{0.405} 
  & \multicolumn{1}{c}{61}  
  & \multicolumn{1}{c}{0.194} \\

  \multicolumn{1}{l|}{StAR~\cite{WWW2021_StAR}}
  & \multicolumn{1}{c}{0.176} 
  & \multicolumn{1}{c}{0.307} 
  & \multicolumn{1}{c}{0.493} 
  & \multicolumn{1}{c}{79}
  & \multicolumn{1}{c}{0.280} \\
    
    
   
\midrule 
  \multicolumn{6}{c}{\textit{Multi-Modal Approaches}} \\
\midrule
     
  
  

  \multicolumn{1}{l|}{TransAE~\cite{IJCNN2019_TransAE}} 
  & \multicolumn{1}{c}{0.274} 
  & \multicolumn{1}{c}{0.489} 
  & \multicolumn{1}{c}{0.715} 
  & \multicolumn{1}{c}{36}
  & \multicolumn{1}{c}{0.421} \\

  \multicolumn{1}{l|}{RSME~\cite{MM2021_RSME}} 
  & \multicolumn{1}{c}{0.485} 
  & \multicolumn{1}{c}{0.687} 
  & \multicolumn{1}{c}{\textbf{0.838}} 
  & \multicolumn{1}{c}{72}
  & \multicolumn{1}{c}{0.607} \\
    
  \multicolumn{1}{l|}{MKGformer~\cite{SIGIR2022_MKGformer}} 
  & \multicolumn{1}{c}{0.448} 
  & \multicolumn{1}{c}{0.651} 
  & \multicolumn{1}{c}{0.822} 
  & \multicolumn{1}{c}{\textbf{23}}
  & \multicolumn{1}{c}{0.575} \\
   
   
  
   
   

\bottomrule

\end{tabular}
    

\end{table*}

\begin{table*}[!htbp]

\centering
\small

\caption{
    \label{tab:openbg500}
    Results of link prediction on \textbf{OpenBG500} and \textbf{OpenBG500-L}. 
    The bold numbers denote the best results.
}

\begin{tabular}{l | c c c c c | c c c c c}
\toprule
\multicolumn{1}{l|}{\multirow{3}*{Model}} 
& \multicolumn{5}{c|}{\bf OpenBG500} & \multicolumn{5}{c}{\bf OpenBG500-L} \\
\cmidrule(lr){2-6} \cmidrule(lr){7-11} 
& Hits@1$\uparrow$ & Hits@3 $\uparrow$ & Hits@10$\uparrow$ & MR$\downarrow$ & MRR$\uparrow$ & Hits@1$\uparrow$ & Hits@3 $\uparrow$ & Hits@10$\uparrow$ & MR$\downarrow$ & MRR$\uparrow$ \\ 
\midrule


TransE~\cite{NeurIPS2013_TransE} 
& 0.207 & 0.340 & 0.513 & 5381 & 0.304  
& \textbf{0.314} & \textbf{0.583}	& \textbf{0.820} & 888 & \textbf{0.482} \\ 

TransH~\cite{AAAI2014_TransH} 
& 0.143 & 0.402 & 0.569 & 6501 & 0.296 
& 0.247	& 0.569	& 0.813	& 1157 & 0.441 \\

TransD~\cite{ACL2015_TransD} 
& 0.146 & 0.411 & 0.576 & 6129 & 0.302 
& 0.279	& 0.575	& \textbf{0.820} & \textbf{858}	& 0.461 \\

DistMult~\cite{ICLR2015_DistMult} 
& 0.068 & 0.131 & 0.255 & 5709 & 0.129
& 0.012	& 0.147	& 0.299 & 3065 & 0.108 \\

ComplEx~\cite{ICML2016_ComplEx}  
& 0.081 & 0.187 & 0.313 & 6393 & 0.156 
& 0.088 & 0.195	& 0.300 & 4569 & 0.165 \\

TuckER~\cite{ACL2019_TuckER} 
& \textbf{0.428} & \textbf{0.615} & \textbf{0.735} & 2573 & \textbf{0.541} 
& - & -	& - & -	& - \\




KG-BERT~\cite{2019_KG-BERT} 
&  0.071 & 0.145 & 0.262 & \textbf{401} & 0.138
& - & - & - & - & - \\


GenKGC~\cite{WWW-Poster2022_GenKGC}
& 0.203 & 0.280 & 0.351 & - & - 
& - & - & - & - & - \\

\bottomrule

\end{tabular}


\end{table*}




\subsection{Baselines and Settings}

To evaluate the quality of {\ours} benchmarks, we implement link prediction experiments on several baselines regarding their varying scales and settings.

(1) \emph{Single-modal KG Embedding Approaches}\footnote{Note that due to the large scale of OpenBG500-L, for some baselines requiring excessive computational resources, \eg, TuckER~\cite{ACL2019_TuckER}, we omit part of experiments as only one V100 GPU is available for benchmark experiments.} for OpenBG500 \& OpenBG500-L \& OpenBG-IMG: TransE~\cite{NeurIPS2013_TransE}, TransH~\cite{AAAI2014_TransH}, TransD~\cite{ACL2015_TransD}, DistMult~\cite{ICLR2015_DistMult}, ComplEx~\cite{ICML2016_ComplEx}, TuckER~\cite{ACL2019_TuckER}, and StAR~\cite{WWW2021_StAR} leverage the KG structure for link prediction.
KG-BERT~\cite{2019_KG-BERT} utilizes the description text in KG embedding with the pre-trained language model.
GenKGC~\cite{WWW-Poster2022_GenKGC} converts KG embedding to sequence-to-sequence generation task with the pre-trained language model.

(2) \emph{Multimodal KG Embedding Approaches} for OpenBG-IMG only: TransAE~\cite{IJCNN2019_TransAE} uses a multimodal auto-encoder based on TransE for a unified representation of textual and visual input. 
RSME~\cite{MM2021_RSME} designs a filter gate and a forget gate to enhance visual information learning. 
MKGformer~\cite{SIGIR2022_MKGformer} utilizes a hybrid transformer architecture with multi-level fusion for better multimodal entity representation. 

\emph{Implementation Details of Baselines.} 
We utilize top-K hit rate (Hits@K, K = 1, 3, 10), Mean Rank (MR), and Mean Reciprocal Ranking (MRR) as evaluation metrics.
Note that OpenBG-IMG is in the multimodal setting with both texts and images, while OpenBG500 and OpenBG500-L follow vanilla settings. 
We conduct experiments on 1$\times$V100 GPU (32G NVIDIA TESLA) in PyTorch 1.8.1 environment. 
With regard to settings of the training process on different baselines, we respectively use AdaGrad and SGD optimizer, with epochs of 5; 100; 200; 500; 1,000. The dimension of embedding is 200, the batch size is set to 32; 100; 200; 500; 1000; 1500, and learning rate is 1.0; 0.5; 5e-4; 3e-5; 1e-5.

\subsection{Overall Performance Comparison}

We have conducted both single-modal and multimodal link prediction on {\ours} benchmarks. 
Link prediction in this paper aims to identify missing connections in a KG based on the features of existing triples. For example, given an incomplete triple $(h, r, ?)$ with head entity $h$ and relation $r$, we aim to predict a tail entity $t$ to complete the triple. $h$ and $t$ can both be of the text modality (single-modal) or respectively be of the text and image modality (multimodal). 

(1) \emph{Single-modal Link Prediction}. 
The experimental results in Table~\ref{tab:openbg-img} and Table~\ref{tab:openbg500} illustrate that translational distance models (TransE, TransH, \etc.) greatly outperform normal bilinear models (DistMult, ComplEx, \etc.).
We notice that TuckER can archive higher Hits@K and MRR scores due to the powerful representation ability of tucker decomposition. 
However, TuckER obtains the worst score on the MR metric in Table~\ref{tab:openbg-img}; we argue that the baseline performance may vary among different instances. 
We further observe that baselines based on textual embeddings such as KG-BERT and StAR are not competitive here.
For OpenBG500-L, we notice that  vanilla TransE can yield better performance than most of the sophisticated baselines, which indicates that more works should be investigated for large-scale KG representation learning.  

(2) \emph{Multimodal Link Prediction}. 
The experimental results in Table~\ref{tab:openbg-img} show that RSME achieves the best performance except on MR metric, indicating its advanced multimodal information fusion ability.
MKGformer can yield better results on MR metric but achieve comparable performance on the other metrics compared with RSME.

\subsection{Distribution and Maintenance}

{\ours} benchmarks\footnotemark[3] have already been released publicly together with an online competition\footnotemark[4].
Up to now, more than \textbf{4,700} researchers have applied {\ours} benchmarks, and over \textbf{3,000} teams have signed up for the competition. 
We will continue to maintain and expand the benchmarks, and plan to enrich them with more new datasets.


\section{{\ours} Pre-Training: KG-Enhanced Application}
\label{sec:application}

\subsection{Overview}

To further unleash the potential of {\ours}, we pre-train the large model enhanced with {\ours} and apply it to e-commerce downstream tasks requiring domain knowledge about items\footnote{Note that an item (\begin{CJK*}{UTF8}{gbsn}商品\end{CJK*}) is regarded as an instance of a product (\begin{CJK*}{UTF8}{gbsn}产品\end{CJK*}), and a product is a standardized expression of a set of identical items, \eg, a product of iPhone 14 Pro comprises many items sold by different retailers.}. 

\emph{Data for Pre-Training.}
The supervised data comprises {\ours} and annotated item titles and reviews, which can be denoted as \emph{label-sample} pairs, such as \emph{product-category; item-product; item-title; item-image; item-triple; short title-long title; item-review; triple-review.}
For texts and KG, we normalize these data in different modals into unified textual expressions with artificially constructed discrete prompts, and obtain almost 100 million supervised text pairs, denoted by $\mathcal{X}_{sup} = (\mathcal{X}_{src}, \mathcal{X}_{tgt})$. 
In addition, we also collect about 140GB of unsupervised texts in e-commerce scenarios, such as user reviews and product descriptions, denoted by $\mathcal{X}_{uns}$. 
We then averagely combine supervised text pairs and unsupervised e-commerce texts together as the input text tokens, denoted by $\mathcal{X} = [\mathcal{X}_{sup}, \mathcal{X}_{uns}]$. Text tokens $\mathcal{X}$ along with visual tokens $\mathcal{V}$ constitute the model input, denoted by $\{ \mathcal{X}, \mathcal{V} \}$. 

\begin{figure}[!htbp]
  \centering
  \includegraphics[width=0.99\linewidth]{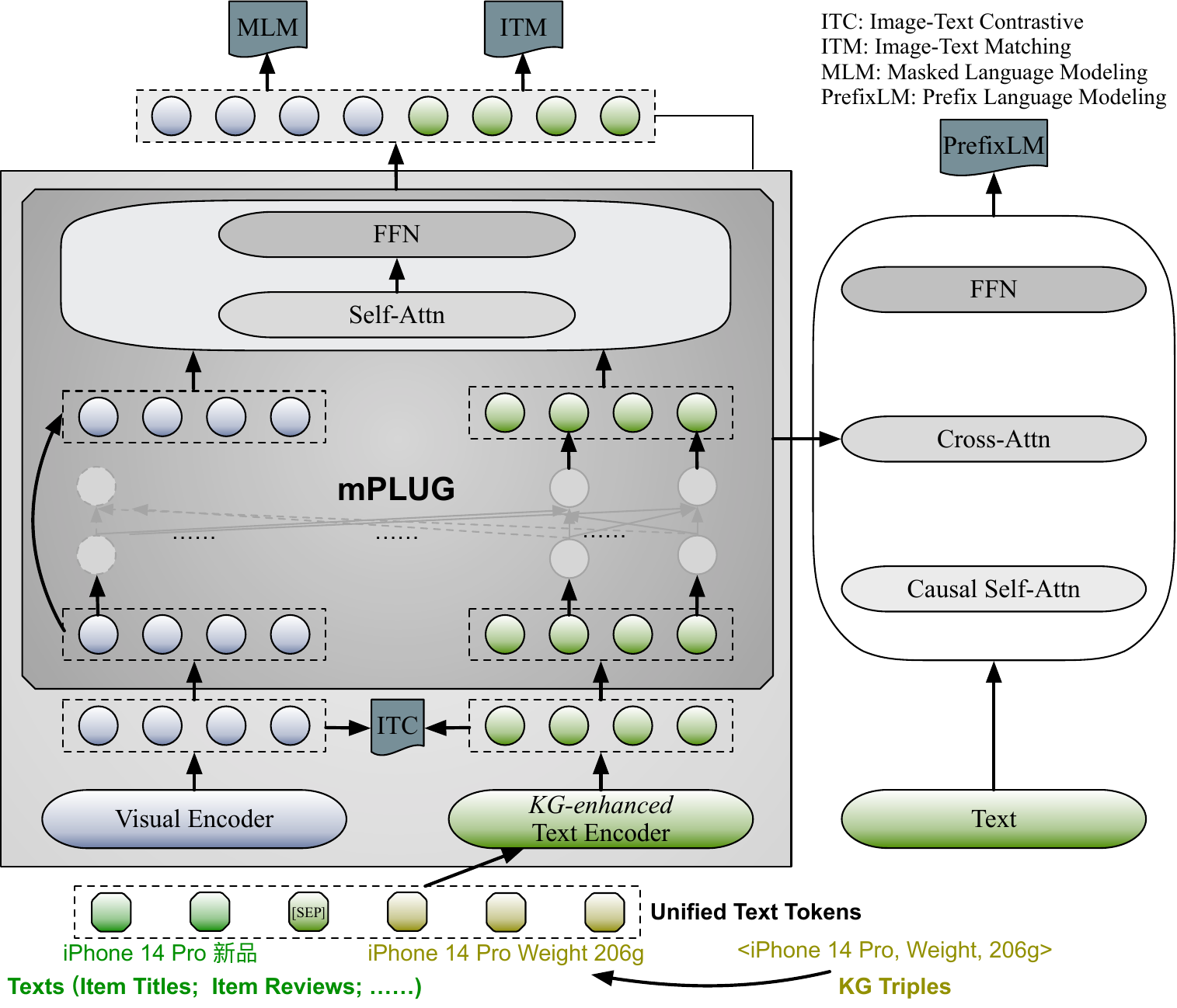}
  \vspace{-3mm}
  \caption{A illustration of the large model pre-training enhanced with {\ours}.  
  \label{fig:exp_model} }
  \vspace{-3mm}
\end{figure}

\emph{Model for Pre-training.}
We illustrate the framework of the large model pre-training enhanced with {\ours} in Figure~\ref{fig:exp_model}. As seen, our pre-training model adopts a vision-language foundation model mPLUG~\cite{EMNLP2022_mPLUG} as the backbone, which is pre-trained end-to-end on large-scale image-text pairs with both discriminative and generative objectives. We revise the text encoder of vanilla mPLUG into a \emph{KG-enhanced text encoder}, where the texts and KG triples are converted into unified text tokens as input. 
We utilize cross entropy as the loss function for supervised data, similar to the multi-task pre-training of ExT5~\cite{ICLR2022_ExT5}; and we adopt the span-denoising objective~\cite{JMLR2020_T5} for unsupervised data. 

\emph{Distributed Pre-training for Large Scale.} 
Training the big mPLUG model on large-scale datasets along with {\ours} is extremely challenging.
Similar to primary mPLUG \cite{EMNLP2022_mPLUG}, to implement time-saving computation, we use BF16 precision training, which is a new data type supported by NVIDIA's new Ampere architecture GPU like A100. 
We have trained mPLUG model for 600,000 steps on 14$\times$A100 GPUs in a data-parallel manner, with the batch size of 512 and weight\_decay of 0.02. We use AdamW optimizer, and adopt a linear schedule to the learning rate with warmup of 0.1. 

\emph{Model Fine-tuning.} 
Each downstream task is formatted as a generation task and fine-tuned separately based on the pre-trained model until the loss no longer decreases on the validation set. To prevent label leakage, the data in the validation set of downstream tasks is restricted not to appear in the pre-training period. 

In this section, we will take five representative downstream tasks as examples, to introduce how to utilize {\ours} to enhance pre-trained large models and serve for real-world applications, including: \textbf{category prediction}, \textbf{NER for titles}, \textbf{title summarization}, \textbf{IE for reviews}, and \textbf{salience evaluation}.

\begin{table*}[!htbp]
\centering
\small

\caption{\label{tab:exp_task} 
	Evaluation results of downstream tasks. \protect\\
	(\emph{As training on a large scale is time-consuming, we only present part of experiments at the time of writing.})
}

\resizebox{\linewidth}{!}{
\begin{tabular}{l | p{2.8cm}<{\centering} | p{2.8cm}<{\centering} | p{2.8cm}<{\centering} | p{2.8cm}<{\centering} | p{2.8cm}<{\centering}} 

\toprule
\textbf{Models} 
& \tabincell{c}{ \textbf{Category Prediction} \\ (Accuracy) } 
& \tabincell{c}{ \textbf{NER for Titles} \\ (P $|$ R $|$ F) }
& \tabincell{c}{ \textbf{Title Summarization} \\ (ROUGE-L) }
& \tabincell{c}{ \textbf{IE for Reviews} \\ (P $|$ R $|$ F) } 
& \tabincell{c}{ \textbf{Salience Evaluation} \\ (Accuracy) } 
\\
\midrule

\textbf{RoBERTa$_\text{large}$}~\cite{2019_RoBERTa} 	
& 68.80 & 70.10 $|$ 69.70 $|$ 69.10 & / & / & / \\


\textbf{UIE}~\cite{ACL2022_UIE} 			
& / & 66.35 $|$ 63.70 $|$ 65.00 & / & / & / \\

\textbf{mT5}~\cite{NAACL2021_mT5} 			
& / & / & 70.12 & 83.42 $|$ 83.21 $|$ 83.32 & / \\

\textbf{BERT}~\cite{NAACL2019_BERT}			
& / & / & / & / & 63.34 \\

\textbf{mPLUG$_\text{base}$}~\cite{EMNLP2022_mPLUG} 	
& 73.10 & 69.68 $|$ 65.98 $|$ 67.78 & 71.82 & 82.60 $|$ 83.06 $|$ 82.83 & 66.45 \\

\midrule

\textbf{mPLUG$_\text{base\textcolor{red}{+KG}}$}~\cite{EMNLP2022_mPLUG} 	
& 74.48 & 73.97 $|$ 72.05 $|$ 73.00 & 72.30 & 83.99 $|$ 83.53 $|$ 83.76 & 69.45 \\




\textbf{mPLUG$_\text{large\textcolor{red}{+KG}}$}~\cite{EMNLP2022_mPLUG} & 74.60 & 75.09 $|$ 72.52 $|$ 73.79 & 78.29 & 84.87 $|$ 84.95 $|$ 84.91 & 69.87 \\



\bottomrule
\end{tabular}
}
\end{table*}

\subsection{Category Prediction}

(1) \emph{Task Definition}: 
Category prediction means link prediction in a KG specifically for item categories. 
Given an incomplete triple ($e$, \texttt{rdfs:subClassOf}, $?$) in which $e$ denotes an item, we require to find the missing tail entity. 
For example, given an item entitled ‘‘Northeast Wuchang Rice Daohuaxiang No.2 (\begin{CJK*}{UTF8}{gbsn}东北五常大米稻花香2号\end{CJK*})'', the model should link the leaf category ‘‘Rice (\begin{CJK*}{UTF8}{gbsn}大米\end{CJK*})'' to it with a relation of \texttt{rdfs:subClassOf}. 

(2) \emph{Baseline Models}: 
We select RoBERTa$_\text{large}$~\cite{2019_RoBERTa} as a baseline model to compare with other methods based on mPLUG. RoBERTa$_\text{large}$~\cite{2019_RoBERTa} is an improved replication of BERT \cite{NAACL2019_BERT} and achieves promising performance in many downstream tasks. Note that the RoBERTa$_\text{large}$ adopted in this task is pre-trained with corpus in general domains and excluding data in the business domain. 

(3) \emph{Dataset Details}: The dataset for category prediction includes a training set of 10 million triples which is generated from crowd-sourced annotations, and a validation set of 210 thousand triples which is created by domain experts. The total number of categories is 13,000 (referring to the entire e-commerce platforms, thus containing more categories than {\ours}). 
For low-resource scenarios, we randomly sample categories with 1 shot and 5 shots respectively. 

(4) \emph{Experimental Analysis}: 
We evaluate category prediction task with the metric of accuracy, and present experimental results in Table~\ref{tab:exp_task}. 
As seen, on the category prediction task, mPLUG$_{base}$ with both discriminative and generative objects is able to achieve better results than the discriminative model, \ie, RoBERTa$_{large}$, and the enhancement with KG also brings about 1.38\% improvement. However, increasing the model capacity (comparing mPLUG$_\text{large\textcolor{red}{+KG}}$ to mPLUG$_\text{base\textcolor{red}{+KG}}$) only achieves 0.12\% outperformance, and we estimate that the phenomenon is probably due to inevitable noises in the training set.

\begin{table}[H]
\centering
\small

\caption{\label{tab:exp_tasks_low_cp} 
	Evaluation results (Accuracy) of low-resource category prediction. 
}

\begin{tabular}{l c c}
\toprule
\textbf{Model} & \textbf{1-Shot} & \textbf{5-Shot} \\
\midrule

\textbf{RoBERTa$_\text{large}$}~\cite{2019_RoBERTa} & 24.16 & 68.73 \\

\textbf{RoBERTa$_\text{base\textcolor{red}{+KG}}$}~\cite{2019_RoBERTa} & 35.74 & 68.99 \\

\textbf{mPLUG$_\text{base}$}~\cite{EMNLP2022_mPLUG} & 37.88 & 67.17 \\

\textbf{mPLUG$_\text{base\textcolor{red}{+KG}}$}~\cite{EMNLP2022_mPLUG} & 48.94 & 70.18 \\

\textbf{mPLUG$_\text{large\textcolor{red}{+KG}}$}~\cite{EMNLP2022_mPLUG} & 57.68 & 71.57 \\

\bottomrule
\end{tabular}

\end{table}

We also evaluate the performance of category prediction in low-resource scenarios, and the results are shown in Table~\ref{tab:exp_tasks_low_cp}. 
As seen, \text{mPLUG$_\text{base\textcolor{red}{+KG}}$} with KG enhanced achieves the outperformance of 11.06\% than mPLUG$_\text{base}$ in the 1-shot setting, 3.01\% in the 5-shot setting. RoBERTa$_\text{base\textcolor{red}{+KG}}$ enhanced with {\ours} even exceeds \text{RoBERTa${_\text{large}}$} by 11.58\% in the 1-shot setting, and 0.26\% in the 5-shot setting. 
Thus, we can conclude that enhance the model with {\ours} can significantly improves performance in few-shot category prediction tasks, and the more deficient data is, the more advantageous to integrate models with KG. 
Besides, increasing the capacity of the model can also helpful, since mPLUG$_\text{large\textcolor{red}{+KG}}$ achieves 8.74\% and 1.39\% outperformance than mPLUG$_\text{base\textcolor{red}{+KG}}$ in the 1-shot and 5-shot setting, respectively.

\subsection{NER for Titles}

(1) \emph{Task Definition}: 
NER (Named Entity Recognition) for titles aims to recognize the properties and corresponding values of items, with the input of item titles. 
For example, given an item entitled ‘‘Zero-fat Konjac Noodles 100g*3 (\begin{CJK*}{UTF8}{gbsn}零脂魔芋面条100克*3袋\end{CJK*})'', the model should find out the property-value pairs, including 
[Nutrients (\begin{CJK*}{UTF8}{gbsn}营养成分\end{CJK*}): Zero-fat (\begin{CJK*}{UTF8}{gbsn}零脂\end{CJK*})];
[Ingredients (\begin{CJK*}{UTF8}{gbsn}原料成分\end{CJK*}): Konjac (\begin{CJK*}{UTF8}{gbsn}魔芋\end{CJK*})];
[Category (\begin{CJK*}{UTF8}{gbsn}品类\end{CJK*}): Noodles (\begin{CJK*}{UTF8}{gbsn}面条\end{CJK*})];
[Packing Specification (\begin{CJK*}{UTF8}{gbsn}包装规格\end{CJK*}): 100g*3 (\begin{CJK*}{UTF8}{gbsn}100克*3袋\end{CJK*})]. 

(2) \emph{Baseline Models}: 
We select RoBERTa$_\text{large}$~\cite{2019_RoBERTa} and UIE~\cite{ACL2022_UIE} as baselines, thereinto, UIE~\cite{ACL2022_UIE} is a unified text-to-structure generation framework, which can uniformly handle different information extraction tasks such as NER, and the model performance is also promising. 
We also pre-train UIE with corpus in general domains. 

(3) \emph{Dataset Details}: The dataset of NER for titles is generated from crowd-sourced annotations, including 1,500\footnote{To annotate NER datasets in business scenarios is actually a hard work, and we will continue to enrich the ongoing annotations.} samples covering 167 entity types in total. The training set and validation set is randomly splitted with the ratio of 8:2. 
For low-resource scenarios, we randomly sample entity types with 1 shot and 5 shots respectively.

(4) \emph{Experimental Analysis}: 
We evaluate NER for titles with the metric of precision (P), recall (R), F1 score (F), and present experimental results in Table~\ref{tab:exp_task}. 
As seen, mPLUG${_\text{base\textcolor{red}{+KG}}}$ outperforms RoBERTa$_\text{large}$ on P, R, F, while mPLUG${_\text{base}}$ performs even worse than RoBERTa$_\text{large}$, indicating the significant advantage of introducing {\ours} to large models. The reason may relate to the great correlation between KG and NER tasks, and titles often contain rich attribute information which are also expressed in KG triples. For example, the item title ‘‘Zero-fat Konjac Noodles 100g*3 (\begin{CJK*}{UTF8}{gbsn}零脂魔芋面条100克*3袋\end{CJK*})'' implies the triple of $\langle$e, Nutrients (\begin{CJK*}{UTF8}{gbsn}营养成分\end{CJK*}), Zero-fat (\begin{CJK*}{UTF8}{gbsn}零脂\end{CJK*})$\rangle$ already existing in KG, where $e$ denotes the corresponding product of the item. 

We also evaluate the performance (F1 score) of NER for titles in low-resource scenarios, and the results are shown in Table~\ref{tab:exp_tasks_low_ner}. 
As seen, in 1-shot setting, mPLUG$_\text{base}$ performs worse than UIE, and mPLUG$_\text{base\textcolor{red}{+KG}}$ enhanced with KG exceeds UIE, demonstrating that KG is significant in low-resource scenarios. Furthermore, both mPLUG$_\text{base}$ and mPLUG$_\text{base\textcolor{red}{+KG}}$ performs worse than RoBERTa$_\text{base\textcolor{red}{+KG}}$ in both 1-shot setting and 5-shot settings, and only mPLUG$_\text{large\textcolor{red}{+KG}}$ obtains outperformance, showing that the model size is also important in low-resource scenarios. 

\begin{table}[!htbp]
\centering
\small


\caption{\label{tab:exp_tasks_low_ner} 
	Evaluation results (F1) of low-resource NER for titles. 
}

\begin{tabular}{l c c}
\toprule
\textbf{Model} & \textbf{1-Shot} & \textbf{5-Shot} \\
\midrule

\textbf{UIE}~\cite{ACL2022_UIE} & 57.20 & 66.80 \\

\textbf{RoBERTa$_\text{base\textcolor{red}{+KG}}$}~\cite{2019_RoBERTa} & 59.60 & 67.90 \\

\textbf{mPLUG$_\text{base}$}~\cite{EMNLP2022_mPLUG} & 40.51 & 50.96 \\

\textbf{mPLUG$_\text{base\textcolor{red}{+KG}}$}~\cite{EMNLP2022_mPLUG} & 57.84 & 61.55 \\

\textbf{mPLUG$_\text{large\textcolor{red}{+KG}}$}~\cite{EMNLP2022_mPLUG} & 62.57 & 70.41 \\

\bottomrule
\end{tabular}

\end{table}


\subsection{Title Summarization}

(1) \emph{Task Definition}: 
Title summarization is designed to summarize the expatiatory titles of items (as item titles contain redundant information in most cases), similar to text summarization. 
For example, given the item title of ‘‘Lagogo Lagu Valley 2018 Summer New Women's Word-neck Short-sleeved Floral Skirt Dress Beach Skirt Long Skirt Tide (\begin{CJK*}{UTF8}{gbsn}Lagogo拉谷谷2018夏新款女一字领短袖碎花裙连衣裙沙滩裙长裙潮\end{CJK*})'', the model can simplify it to ‘‘Lagogo One-Line Neck Short Sleeve Floral Dress (\begin{CJK*}{UTF8}{gbsn}Lagogo一字领短袖碎花连衣裙\end{CJK*})''. 

(2) \emph{Baseline Models}: 
We select mT5~\cite{NAACL2021_mT5} as our baseline model for title summarization. 
mT5~\cite{NAACL2021_mT5} is a multilingual variant of T5~\cite{JMLR2020_T5}, the text-to-text transfer transformer, and demonstrates great performance on many multilingual benchmarks. 
We also pre-train mT5 with corpus in general domains. 

(3) \emph{Dataset Details}: The dataset for title summarization is built by domain experts with a total of 7 million samples. The training set and validation set is randomly splitted with the ratio of 8:2. 

(4) \emph{Experimental Analysis}: 
We evaluate title summarization with the metric of ROUGE-L, and present experimental results in Table~\ref{tab:exp_task}. 
As seen, mPLUG${_\text{base}}$ already excels mT5, and mPLUG$_\text{base\textcolor{red}{+KG}}$ enhanced with KG even more predominant. 
The probable reasons for the outperformance are mainly comprises two aspects:
The first one is that we pre-train mPLUG$_\text{base}$ with Chinese corpus, which can achieve better performance in Chinese datasets compared to mT5;
the second one is that KG with attribute triples can help to identify phrases containing key attributes of items, so that making text summarization more meaningful. 
Besides, increasing the model capacity can also bring more significant improvement than introducing KG information in title summarization.

\subsection{IE for Reviews}

(1) \emph{Task Definition}: 
IE (Information Extraction) for reviews aims at extracting structured information which describes and evaluates the item from the customer reviews. 
For example, given the item review of ‘‘The quality of the cushion is good, I bought it for my dad, the size is right, he likes it very much (\begin{CJK*}{UTF8}{gbsn}坐垫质量不错，给老爸买的，大小合适，他很喜欢\end{CJK*})'', the model should figure out some triples, such as $\langle$cushion (\begin{CJK*}{UTF8}{gbsn}坐垫\end{CJK*}), quality (\begin{CJK*}{UTF8}{gbsn}质量\end{CJK*}), nice (\begin{CJK*}{UTF8}{gbsn}不错\end{CJK*})$\rangle$ and $\langle$cushion (\begin{CJK*}{UTF8}{gbsn}坐垫\end{CJK*}), size (\begin{CJK*}{UTF8}{gbsn}大小\end{CJK*}), suitable (\begin{CJK*}{UTF8}{gbsn}合适\end{CJK*})$\rangle$. 

(2) \emph{Baseline Models}: 
We also select mT5~\cite{NAACL2021_mT5} as our baseline model to implement IE for item reviews. We covert IE into a generative task and also pre-train mT5 model with general-domain corpus. 

(3) \emph{Dataset Details}: The dataset of IE for item reviews is created by crowd-sourced annotations, including 20,000 samples in total. The training set and validation set is randomly splitted with the ratio of 8:2.

(4) \emph{Experimental Analysis}:  
We evaluate IE for reviews with the metric of precision (P), recall (R), F1 score (F), and present experimental results in Table~\ref{tab:exp_task}. 
As seen, mPLUG$_\text{base}$ underperforms than mT5, while when enhanced with KG, 
mPLUG$_\text{base\textcolor{red}{+KG}}$ exceeds mT5. 
The possible reasons is that taxonomy knowledge defined in {\ours} can help the model to distinguish the types of various entities in the texts, and attribute knowledge contains in {\ours} can help to implement attribute extraction of items.

\subsection{Salience Evaluation}

(1) \emph{Task Definition}: 
Salience evaluation for commonsense knowledge in business scenarios is defined at \cite{EMNLP2022_KnowledgeSalienceEval}, where salience~\cite{AKBC2020_Commonsense} reflects if a statement (triple) is characteristic enough, meaning that the relation in the triple can be regarded as a key trait of the entity. 
Given a triple $t = \langle e_h, r, e_t \rangle$, the model should to estimate if $t$ is salient (score to 1) or not (score to 0). 
For example, $\langle$running shoes, \texttt{relatedScene}, running$\rangle$ is salient and should score to 1, while $\langle$shoes, \texttt{relatedScene}, running$\rangle$ is not and should score to 0. 

(2) \emph{Baseline Models}: 
We select BERT~\cite{NAACL2019_BERT} as the baseline model for commonsense knowledge salience evaluation in business scenarios, which stands for bidirectional encoder from transformers and gains satisfactory achievements in many NLP tasks. BERT is also pre-trained with general-domain corpus. 

(3) \emph{Dataset Details}: The dataset for salience evaluation is also created by domain experts, including 23,000 samples in total. The training set and validation set is randomly splitted with the ratio of 8:2.

(4) \emph{Experimental Analysis}:
We evaluate salience of triples with the metric of accuracy, and present experimental results in Table~\ref{tab:exp_task}. 
As seen, vanilla mPLUG (mPLUG$_\text{base}$) and KG enhanced mPLUG models (mPLUG$_\text{base\textcolor{red}{+KG}}$, mPLUG$_\text{large\textcolor{red}{+KG}}$) all excels BERT at accuracy. Moreover, the outperformance gained from KG enhancement is more prominent than from model capacity expansion, as mPLUG$_\text{large\textcolor{red}{+KG}}$ merely exceeds mPLUG$_\text{base\textcolor{red}{+KG}}$ by 0.42\%, and both of them can exceed BERT by more than 7\%. 
This result is instinctive, because salience evaluation is heavily dependent on commonsense knowledge in retail, which is exactly the crucial constituent in {\ours} and mainly expressed by concepts.

\subsection{Other Applications and Online Cases} \label{sec:kg_application_other}

\begin{figure}[!htbp]
  \centering
  \subfigure[Channel of Taobao Foodies.]{
  \label{fig:exp_case1}
  \includegraphics[width=0.46\linewidth]{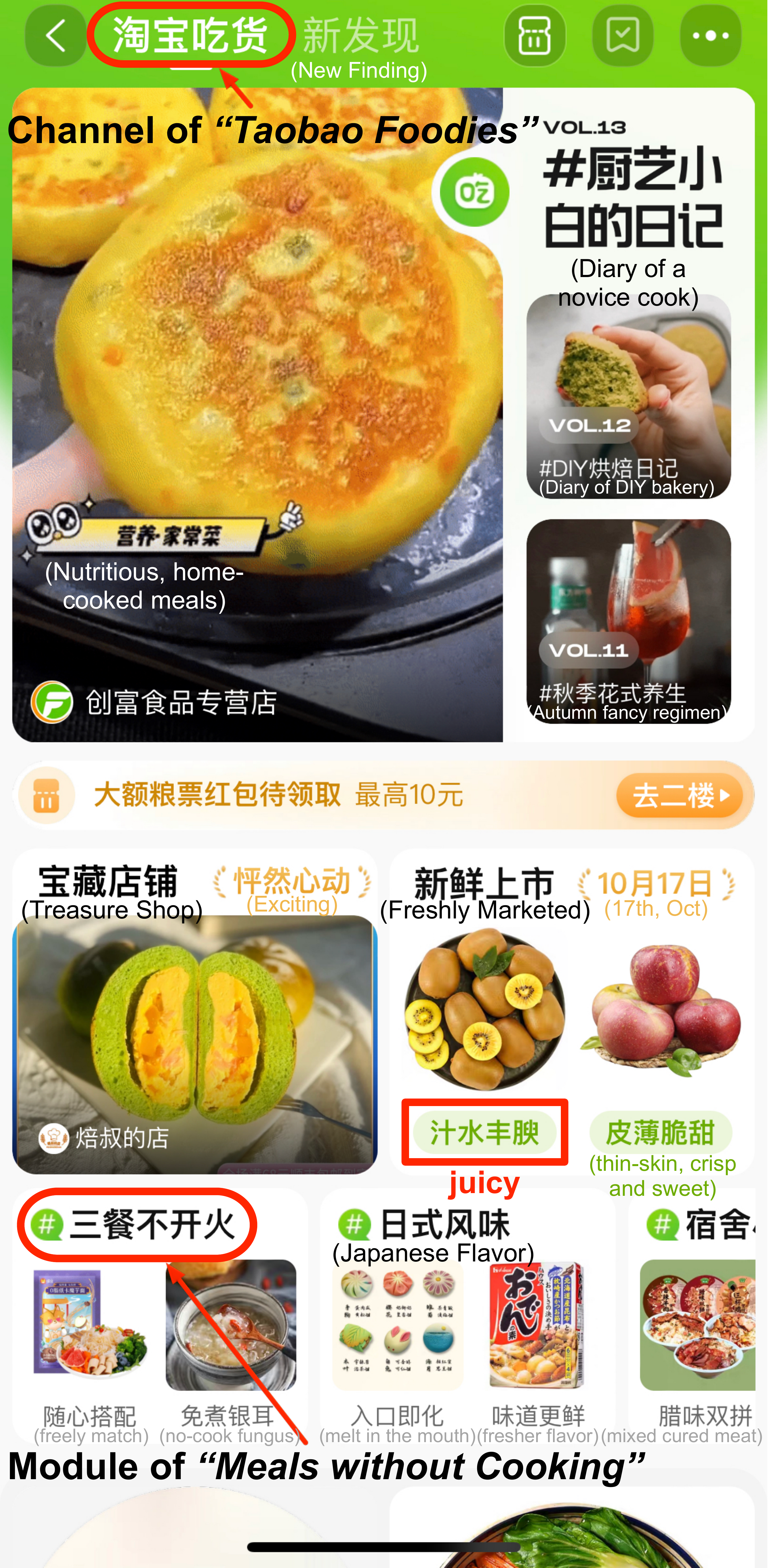}
  }
  \subfigure[Module of Meals without Cooking.]{
  \label{fig:exp_case2}
  \includegraphics[width=0.464\linewidth]{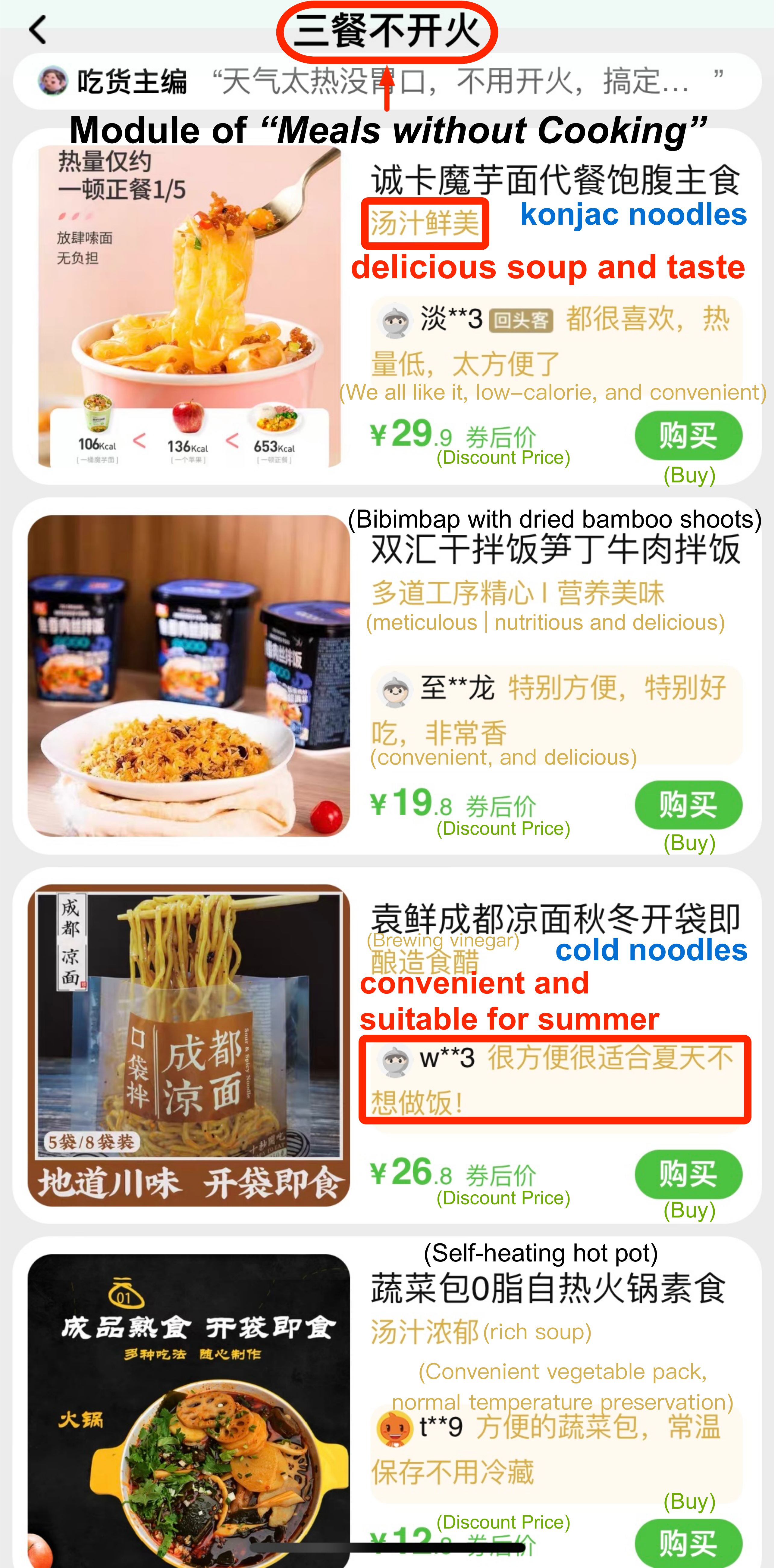}
  }
  \caption{
  \label{fig:cases_online}
     A demonstration of intelligent shopping guide based on pre-trained {\ours}, 
     at the Module of ‘‘Meals without Cooking (\protect\begin{CJK*}{UTF8}{gbsn}三餐不开火\protect\end{CJK*})'', 
     in the Channel of ‘‘Taobao Foodies (\protect\begin{CJK*}{UTF8}{gbsn}淘宝吃货\protect\end{CJK*})'' on Taobao App.
  }
\end{figure}

\emph{Cases on Online Systems.} We deploy the pre-trained large model enhanced with {\ours} on some online systems of Alibaba Group, such as Taobao App (\begin{CJK*}{UTF8}{gbsn}手机淘宝\end{CJK*}). As seen in Figure~\ref{fig:cases_online}, when users enter the Channel of ‘‘Taobao Foodies (\begin{CJK*}{UTF8}{gbsn}淘宝吃货\end{CJK*})'' on Taobao App, they will find some slogans and tips with the item to provide shopping guide, which are summarized from item titles and reviews based on the pre-trained KG-enhanced large model. 
For example, at the Module of ‘‘Meals without Cooking (\begin{CJK*}{UTF8}{gbsn}三餐不开火\end{CJK*})'', an item of ‘‘konjac noodles (\begin{CJK*}{UTF8}{gbsn}魔芋面\end{CJK*})'' is marked as ‘‘delicious soup and taste (\begin{CJK*}{UTF8}{gbsn}汤汁鲜美\end{CJK*})'', and an item of ‘‘cold noodles (\begin{CJK*}{UTF8}{gbsn}凉面\end{CJK*})'' presents a short review of ‘‘convenient and suitable for summer (\begin{CJK*}{UTF8}{gbsn}方便，适合夏天\end{CJK*})''. These slogans and tips align with items can help users promptly pick out the items that they intend to purchase, making the shopping guide more intelligent and user-friendly. 

Pre-trained models enhanced with {\ours} can also be applied to numerous real-world applications in business scenarios, including in this paper but not limited to: 

(1) \emph{Item Alignment}. 
The target is to identify different items referring to the same product. 
With {\ours}, items referring to the same product can be recognized more easily based on the product schema, such as the corresponding category and attributes. 
The evaluation metric is Gross Merchandise Volume (GMV) of aligned items, and larger is better. 
GMV of Alibaba e-commerce platforms has promoted about 45\% after deploying the pre-trained KG-enhanced large model.

(2) \emph{Shopping Guide}. 
The target is to guide users to purchase items.
With {\ours}, items can be tagged with various concepts and linked with more enriched information, so that users can better understand items and intend to consume. 
The evaluation metric is Cost Per Mille (CPM), and larger is better. 
CPM of Alibaba e-commerce platforms has promoted about 28.1\% after deploying the pre-trained KG-enhanced large model.

(3) \emph{QA-based Product Recommendation}. 
The target is to recommend items to users in QA scenarios. 
With {\ours}, AliMe (a smart QA robot) \cite{CIKM2017_AliMeAssist} can better understand users' intention and recommend items more precisely. 
The evaluation metric is Click-Through-Rate (CTR), and larger is better. 
CTR of Alibaba e-commerce platforms has promoted about 11\% after deploying the pre-trained KG-enhanced large model.

(4) \emph{Emerging Product Release}. 
The target is to release emerging products. 
With {\ours}, when emerging products release, their attribute information can be automatically filled in via inheriting from the categories, so as to improve efficiency and save time. 
The evaluation metric is duration time of emerging products release, and smaller is better. 
Alibaba e-commerce platforms has reduced about 30\% duration time to release emerging products after deploying the pre-trained KG-enhanced large model.


\section{Related Work}
\label{sec:related_work}

\subsection{Existing Business KGs (not open-source or comprehensive)} 

As business KGs have been pivotal in providing various knowledge-based services to e-commerce platforms, there exists some business KGs at present, but they are not publicly available and comprehensive enough. 

\emph{AmazonKG} has shown several subsets (not open-source) in different ways: 
\emph{AmazonKG}$_{\tiny{TXtract}}$~\cite{ACL2020_TXtract} with a hierarchical taxonomy contains 2 million products from 4,000 categories without images. 
\emph{AmazonKG}$_{\tiny{AutoKnow}}$~\cite{KDD2020_AutoKnow} contains $>$30M products in 11K unique types and $>$1B triples, and is also single-modal. 
\emph{AmazonKG}$_{\tiny{PAM}}$~\cite{KDD2021_PAM} is multimodal and contains 61,308 samples covering 14 product categories. 

Alibaba Group has also introduced some subset KGs: 
\emph{AliCoCo}~\cite{SIGMOD2020_AliCoCo} is a concept KG containing 57,125 types of concepts and 2 types of relations at the time of writing. 
\emph{AliCoCo2}~\cite{KDD2021_AliCoCo2} enriches AliCoCo with commonsense knowledge and contains 163,460 types of concepts and 91 types of relations at writing time. 
\emph{AliMe KG}~\cite{CIKM2020_AliMeKG} contains thousands of items. 
\emph{AliMe MKG}~\cite{CIKM2021_AliMeMKG} enriches AliMe KG with millions of images. 
These business KGs are all well constructed but not directly open-source to the public. 

\emph{WalmartKG}~\cite{WSDM2020_WalmartKG} is derived from \url{grocery.walmart.com}, containing almost 140,000 common grocery products with short textual descriptions excluding images, and not open-source. 
\emph{Atlas}~\cite{SIGIR2020_Atlas} is a multimodal product taxonomy dataset focusing on clothing products, containing thousands of product images along with their corresponding product titles. Atlas is publicly available but only involves one category -- clothing, and consequently not comprehensive enough.

\subsection{Business KG Construction Approaches}

As datasets in business scenarios are large-scale and complex, developing a high quality business KG is still laborious and challenging. Some previous works have focused on business KG construction from different aspects. 

\emph{Taxonomy Construction.} 
\cite{ICDE2022_ProductTaxoExp} proposes to append new concepts into existing taxonomies with an adaptively self-supervised and user behavior-oriented product taxonomy expansion framework. 
\cite{TMIS2020_KGC_via_MT} proposes a new paradigm that categorizes e-commerce products into a multi-level taxonomy tree with thousands of leaf categories based on machine translation. 
\cite{AAAI2021_KGC_TaxonomyComp} formulates a new task of taxonomy completion by discovering both the hypernym and hyponym concepts for a query, and proposes a triplet matching network to resolve this task. 
\cite{ACL2020_TXtract} proposes to organize thousands of product categories in a hierarchical taxonomy with a taxonomy-aware knowledge extraction model named TXtract. 

\emph{Attribute Value Extraction.} 
\cite{KDD2021_PAM} presents PAM model, a unified framework for multimodal attribute value extraction from product profiles. 
\cite{KDD2018_OpenTag} develops a novel deep tagging model named OpenTag to discover missing values of product attributes by leveraging product profiles. 
\cite{ACL2021_AdaTag} proposes AdaTag, utilizing adaptive decoding to handle multi-attribute value extraction from product profiles. 

\cite{KDD2020_AutoKnow} takes both taxonomy construction and attribute value extraction into account, and describes an automatic (self-driving) system: AutoKnow, to automate ontology construction, knowledge enrichment and cleaning for massive products.

\subsection{Business KG Pre-Training and Embedding for Applications}

\emph{Business KG Pre-training.}
\cite{CVPR2022_M5Product} contributes a large-scale multimodal pre-training dataset named M5Product with self-harmonized contrastive learning, which contains more than 6 million multimodal samples from 6,232 categories. 
\cite{ICDE2021_PKGM} proposes a pre-trained KG model named PKGM for a billion-scale product KG (single-modal), to provide knowledge services for e-commerce tasks. 
\cite{MM2021_K3M} proposes K3M model to introduce knowledge modality in multimodal pre-training \wrt modal-interaction of images, texts and structured knowledge. 

\emph{Business KG Embedding.}
\cite{WSDM2020_WalmartKG} presents a self-attention-enhanced distributed KG embedding method with an efficient multi-task training schema, aiming at learning intrinsic product relations. 
\cite{IJCKG2021_E-commerceKGE} discusses some experiences of deploying KG embedding in e-commerce applications, and identifies the necessity of attentive reasoning, explanations and transferable rules in business scenarios. 
\cite{JIST2019_XTransE} proposes an explainable KG embedding method named XTransE for link prediction with lifestyles in e-commerce, and generates explanations for item-lifestyle prediction results.


\section{Discussion and Conclusion}
\label{sec:con_fw}

\subsection{Lessons Learned} 

In this paper, we demonstrate the feasibility of constructing a billion-scale KG in the vertical domain, and outline the process of designing ontologies, constructing multimodal KG, and managing its quality. 
Despite our efforts, this work still has limitations, as the constructed KG still contains numerous erroneous or missing facts due to the noisy raw data in applications. Therefore, more work should be considered for quality management, such as KG completion and refinement. 
Overall, three lessons should be noted when building a business KG. 
(1) First, \emph{coherence with business logic is crucial}. 
Practical applications impose constraints on the relationships in business KGs by defining an ontology that aligns with business logic. 
To build an accurate, unambiguous, easy-to-maintain, and scalable ontology, developers should possess a deep understanding of the business logic and anticipate potential future changes, so that KG construction can be efficient. 
(2) Second, \emph{practicality and minimalism are essential}. 
Since not all knowledge is necessary for downstream tasks and many incorrect facts may exist in the raw corpus, it is important to integrate relevant and valuable knowledge (\eg, images are valuable for product alignment) to business KG, rather than including all available information. 
(3) Third, \emph{quality control is a lifelong work}. 
Ensuring the integrity of new knowledge, assessing quality, and rectifying incorrect facts require constant attention. 
Moreover, it is critical to manage changes to business KG without introducing inconsistencies with existing knowledge.

\subsection{Conclusion}

In this work, we construct and publish {\ours}, an open business knowledge graph derived from a deployed system. 
To the best of our knowledge, it is the most comprehensive business KG available on the Web. 
{\ours} is integrated with fine-grained taxonomy and multimodal facts and enables linking to a variety of external data. 
As an ongoing business KG, {\ours} currently contains more than 2.6 billion triples, with more than 88 million entities covering over 1 million core classes/concepts and 2,681 types of relations. 
We also release the {\ours} benchmark and run up an online competition based on it. 
We report experimental results of KG-centric tasks on the {\ours} benchmark and KG-enhanced tasks on full {\ours}, illustrating the potential advantages of multimodal structured knowledge in business scenarios.
We release all the data resources for the community to promote future work. 

In the future, we will keep developing and improving {\ours}, and hopefully, it can benefit KG construction in other vertical domains, not limited to business.
Interesting future directions include: 
(1) applying {\ours} to other tasks, such as product recommendation, and 
(2) exploring {\ours} for cross-discipline science, such as economics and sociology.

\section*{Acknowledgment}
We would like to express gratitude to the anonymous reviewers for their kind and helpful comments. 
This work was supported by the National Natural Science Foundation of China (No.62206246, 91846204 and U19B2027), Zhejiang Provincial Natural Science Foundation of China (No. LGG22F030011), Ningbo Natural Science Foundation (2021J190), Yongjiang Talent Introduction Programme (2021A-156-G), and NUS-NCS Joint Laboratory (A-0008542-00-00).

\section*{Contributors} 
\textbf{Shumin Deng} from National University of Singapore conducted the whole development of {\ours} when she was a research intern at Alibaba Group, advised the project, suggested tasks, and wrote the paper.

\textbf{Chengming Wang, Zelin Dai, Hehong Chen, Feiyu Xiong, Ming Yan, Qiang Chen} from Alibaba Group helped to construct {\ours}, and developed the KG-enhanced applications, including pre-trained models and conducted downstream tasks.

\textbf{Zhoubo Li} from Zhejiang University conducted the KG-centric applications and will provide consistent maintenance.

\textbf{Mosha Chen} from Alibaba Group maintained the online data platform and organized the competition at Tianchi.

\textbf{Jiaoyan Chen} from The University of Manchester proofread the paper and advised the project.

\textbf{Jeff Z. Pan} from University of Edinburgh proofread the paper and advised the project.

\textbf{Bryan Hooi} from National University of Singapore proofread the paper and advised the project.

\textbf{Ningyu Zhang, Huajun Chen} from Zhejiang University advised the project, suggested tasks, and led the research.


\balance
\bibliography{custom}
\bibliographystyle{IEEEtran}



\end{document}